\documentclass[sigconf]{acmart}

\AtBeginDocument{%
  }

\setcopyright{acmlicensed}
\copyrightyear{2018}
\acmYear{2018}
\acmDOI{XXXXXXX.XXXXXXX}

\usepackage{algorithm,algorithmic}
\usepackage{enumitem}

\acmConference[Conference acronym 'XX]{Make sure to enter the correct
  conference title from your rights confirmation emai}{June 03--05,
  2018}{Woodstock, NY}
\acmISBN{978-1-4503-XXXX-X/18/06}

\begin{document}

\title{SepsisCalc: Integrating Clinical Calculators into Early Sepsis Prediction via Dynamic Temporal Graph Construction}


\author{Changchang Yin}
\email{yin.731@osu.edu}
\affiliation{
  \institution{The Ohio State University} 
  \city{Columbus}
  \state{Ohio}
  \country{USA}
}


\author{Shihan Fu}
\email{sh.fu@northeastern.edu}
\affiliation{
  \institution{Northestern University} 
  \city{Boston}
  \state{Massachusetts}
  \country{USA}
}

\author{Bingsheng Yao}
\email{b.yao@northeastern.edu}
\affiliation{
  \institution{Northestern University} 
  \city{Boston}
  \state{Massachusetts}
  \country{USA}
}

\author{Thai-Hoang Pham}
\email{pham.375@osu.edu}
\affiliation{
  \institution{The Ohio State University} 
  \city{Columbus}
  \state{Ohio}
  \country{USA}
}

\author{Weidan Cao}
\email{weidan.cao@osumc.edu}
\affiliation{
  \institution{The Ohio State University Wexner Medical Center} 
  \city{Columbus}
  \state{Ohio}
  \country{USA}
}

\author{Dakuo Wang}
\email{d.wang@northeastern.edu}
\affiliation{
  \institution{Northestern University} 
  \city{Boston}
  \state{Massachusetts}
  \country{USA}
}

\author{Jeffrey Caterino}
\email{jeffrey.caterino@osumc.edu}
\affiliation{
  \institution{The Ohio State University Wexner Medical Center} 
  \city{Columbus}
  \state{Ohio}
  \country{USA}
}

\author{Ping Zhang}
\email{zhang.10631@osu.edu}
\affiliation{
  \institution{The Ohio State University} 
  \city{Columbus}
  \state{Ohio}
  \country{USA}
}
\authornote{Corresponding Author}

\renewcommand{\shortauthors}{Yin et al.}
\newcommand{\todo}[1]{\textcolor{red}{(TODO: #1)}}
\newcommand{\ours}{\textcolor{black}{SepsisCalc}}
\newcommand{\here}{\textcolor{red}{CY: I am here.}}


\begin{abstract}

Sepsis is an organ dysfunction caused by a deregulated immune response to an infection. Early sepsis prediction and identification allow for timely intervention, leading to improved clinical outcomes. Clinical calculators (\textit{e.g.}, the six-organ dysfunction assessment of SOFA in \autoref{fig:workflow}) play a vital role in sepsis identification within clinicians' workflow, providing evidence-based risk assessments essential for sepsis diagnosis. However, artificial intelligence (AI) sepsis prediction models typically generate a single sepsis risk score without incorporating clinical calculators for assessing organ dysfunctions, making the models less convincing and transparent to clinicians. To bridge the gap, we propose to mimic clinicians’ workflow with a novel framework SepsisCalc to integrate clinical calculators into the predictive model, yielding a clinically transparent and precise model for utilization in clinical settings. Practically, clinical calculators usually combine information from multiple component variables in Electronic Health Records (EHR), and might not be applicable when the variables are (partially) missing. We mitigate this issue by representing EHRs as temporal graphs and integrating a learning module to dynamically add the accurately estimated calculator to the graphs. Experimental results on real-world datasets show that the proposed model outperforms state-of-the-art methods on sepsis prediction tasks. Moreover, we developed a system to identify organ dysfunctions and potential sepsis risks, providing a human-AI interaction tool for deployment, which can help clinicians understand the prediction outputs and prepare timely interventions for the corresponding dysfunctions, paving the way for actionable clinical decision-making support for early intervention.

\end{abstract}



\maketitle

\begin{figure}[]
    \centering
    \includegraphics[width=\linewidth]{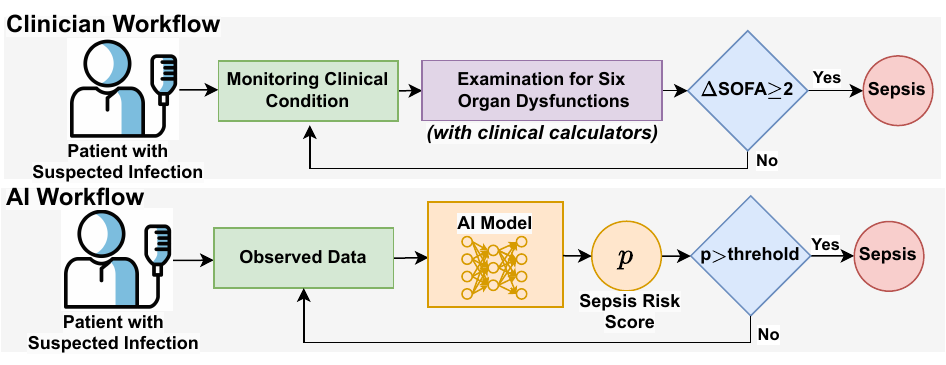} 
    \caption{Workflows of clinicians and AI for sepsis identification. 
    Clinicians examine sepsis by assessing organ dysfunctions with multiple clinical calculators as evidence, while AI workflow only gives an overall sepsis risk score.
    }
    \label{fig:workflow} 
\end{figure}

\section{Introduction}
\label{sec:intro} 

Sepsis, defined as life-threatening organ dysfunction in response to infection, contributes to up to half of all hospital deaths and is associated with more than \$24 billion in annual costs in the United States \cite{liujama}. 
Existing studies \cite{liu2017timing} have shown that sepsis patients may benefit from a 4\% higher chance of survival if diagnosed 1 hour earlier, so developing early sepsis prediction systems can significantly improve clinical outcomes.  
Over the past few decades, the rapid growth in volume and diversity of electronic health records (EHR) has made it possible to adopt state-of-the-art artificial intelligence (AI) methods to early identify patients with sepsis risk.

Deep learning (DL) methods have been widely used for early sepsis risk prediction tasks and have achieved superior performance. However, most existing DL or AI methods are data-driven and fail to incorporate the important and widely used clinical calculators. For example, Sequential Organ Failure Assessment (SOFA)~\cite{vincent1996sofa} serves as an important screening tool for the identification of organ dysfunctions for sepsis in clinicians' workflow, as shown in \autoref{fig:workflow}.
Such a gap in the workflows makes the AI models less convincing and hinders their application to real-world clinical scenarios. 


To bridge the gap, we propose to mimic clinicians' workflow by incorporating clinical calculators into automatic sepsis prediction tasks. 
However, directly feeding calculators into DL models presents a significant challenge:
the calculators usually combine information from multiple component variables (\textit{e.g.}, SOFA has 6 component scores with 12 clinical variables, as shown in \autoref{tab:sofa}), while the variables might have high missing rates (\textit{e.g.}, the missing rates $>60\%$ for most variables in \autoref{tab:all-variables}), making calculators sometimes not applicable. An intuitive method to handle missing values is imputation. However, when the missing rate is high, the imputation models~\cite{3dmice,detroit,yin2020identifying} become less accurate and introduce more bias, which could be harmful for downstream sepsis prediction tasks.    
 
To address the challenges, we propose a novel early \textbf{Sepsis} prediction model with clinical \textbf{Calc}ulators (\textbf{\ours}). 
For each patient, we first construct a temporal graph containing all the observed variables (including demographics, vital signs, lab tests, procedures, and medications). Then we use the graph to estimate the clinical calculators that can summarize six organ function statuses, which are dynamically added to the patient temporal graph, as shown in \autoref{fig:graph_construction}(C). We only include the accurately estimated calculators, filtering out those with all components unobserved due to low confidence.
Finally, we use a graph neural network (GNN) to extract the features of the dynamic temporal heterogeneous graph and make predictions for both organ dysfunctions and sepsis risks.

To demonstrate the effectiveness of our \ours, we conducted extensive experiments on real-world datasets MIMIC-III~\cite{mimic}, AmsterdamUMCdb~\cite{amsterdamumcdb}, and one proprietary dataset extracted from Ohio State University Wexner Medical Center (OSUWMC). Experimental results show that the proposed model outperforms state-of-the-art methods on the early sepsis prediction tasks. Moreover, we developed a system integrated into OSUWMC EHR system to identify organ dysfunctions and potential sepsis risks, providing a human-AI interaction tool for deployment and paving the way for actionable clinical decision-making support for early intervention.

We summarize our contributions as follows:
\begin{itemize} 
    \item We propose a novel sepsis prediction model \ours, which can represent patients' EHR data as dynamic temporal graphs, and effectively extract temporal information, clinical event interaction, and organ dysfunction information from the EHR data.
    \item We incorporate the widely-used and well-validated clinical calculators by dynamically generating the calculator nodes, which can significantly improve prediction performance and make the model more stable and convincing to clinicians.
    \item We conducted extensive experiments on various real-world datasets and the experimental results show that the proposed models outperform state-of-the-art methods on sepsis prediction tasks, demonstrating the effectiveness of the proposed \ours.
    \item We developed a system integrated into EHR system, allowing clinicians to easily use and effectively interact with the models.
\end{itemize}

\section{Related Work}

In this section, we briefly review the existing work related to sepsis prediction systems.  

\noindent
\textbf{Sepsis Screening Tools.} 
Sepsis is a heterogeneous clinical syndrome that is the leading cause of mortality in hospital intensive care units (ICUs) \cite{sepsis3,yin2020identifying}. 
Early prediction and diagnosis may allow for timely treatment and lead to more targeted clinical interventions. 
Screening tools have been used clinically to recognize sepsis, including qSOFA \cite{sepsis3}, MEWS \cite{subbe2001validation}, NEWS \cite{smith2013ability}, and SIRS \cite{bone1992definitions}. However, those tools were designed to screen existing symptoms as opposed to explicitly early predicting sepsis before its onset. 

\begin{figure}
    \centering 
    \includegraphics[width=\linewidth]{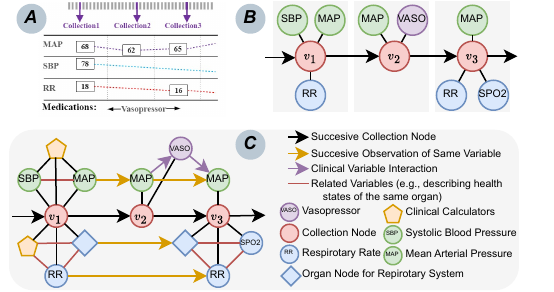}
    \caption{Different EHR representation methods. (A) An example of sequential representation. (B) Example of graph representation with temporal information of clinical observations. (C) The proposed dynamic temporal graph representation with clinical event interaction and clinical calculators. Note that only partial calculator and organ nodes and edges are plotted for graph illustration in subfigure C.}
    \label{fig:graph_construction}
\end{figure}

\noindent
\textbf{Sequence-Based Models.}
With recent advances, deep learning methods have shown great potential for accurate sepsis prediction~\cite{islam2019prediction, reyna2019early,zhang2021interpretable,kamal2020interpretable}. Most of the studies represent patients' EHRs as the sequence of observations (see \autoref{fig:graph_construction}(A)).
Although the methods achieved superior performance, they face a critical challenge due to the data representation. 
The models need to completely observe a list of variables (including vital signs and lab tests), while many variables are missing in real-world data. Existing studies~\cite{zhang2021interpretable,kamal2020interpretable,islam2019prediction} usually impute the missing values before the prediction, raising a new problem that the sepsis prediction models will heavily rely on the imputation methods. The imputation bias would also be propagated to downstream prediction models.


\noindent
\textbf{Graph-Based Models.} Graph representation can naturally handle the missing variables without imputation. However, most existing graph-based models~\cite{gram,kame,yin2019domain,cgl,KerPrint} are designed to model longitudinal sparse EHRs (\textit{e.g.}, diagnosis codes and procedures with binary values) for chronic disease prediction (\textit{e.g.}, heart failure and COPD), the studies on dense float-value variables (\textit{e.g.}, vital signs and lab tests with multiple observations in the same visit) for acute diseases (\textit{e.g.}, sepsis) are still limited. 
Existing works~\cite{chowdhury2023predicting,liu2020hybrid} construct a temporal graph with the observed variables (see \autoref{fig:graph_construction}(B)), eliminating the need for additional imputation methods and avoiding potential imputation bias. However, they still suffer from two limitations: (i) Lack of consideration of clinical event interaction (\textit{e.g.}, vasopressor is used due to the extremely low MAP in \autoref{fig:graph_construction}(B)); (ii) Lack of consideration of clinical calculators that provide clinicians with evidence-based risk assessments essential for accurate diagnosis and prognostic evaluation~\cite{jin2024agentmd}. These limitations make these models unconvincing to clinicians, hindering their application in real-world clinical scenarios.

To address the challenges, we propose to dynamically construct temporal heterogeneous graphs (see \autoref{fig:graph_construction}(C)) that (i) contain temporal relations between observations, (ii) include clinical event interaction, and (iii) estimate and integrate clinical calculators into graphs, to mimic clinicians' workflow. Based on the dynamic temporal graph, we adopt graph neural networks to predict sepsis risk scores with potential organ dysfunction, paving the way for actionable clinical decision-making support for early intervention.




 
\begin{figure}
    \centering
    \includegraphics[width=\linewidth]{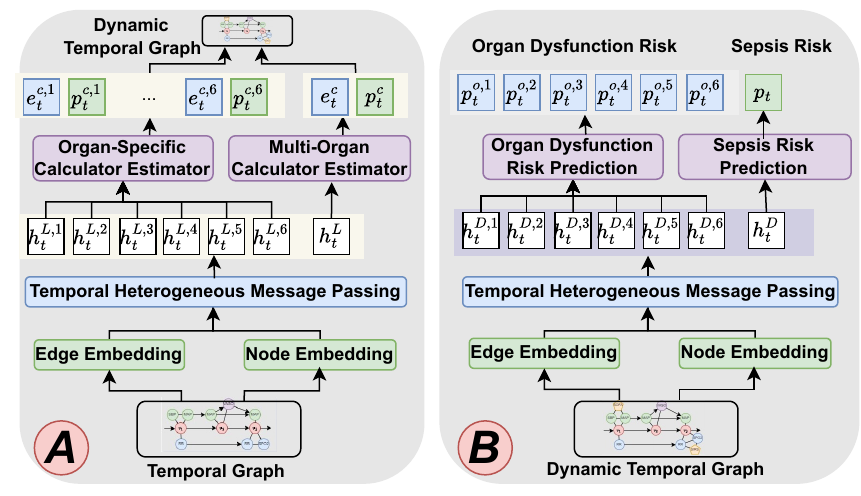}
    \caption{Framework of \ours. (A) Dynamic temporal graph construction. (B) Sepsis prediction framework.}
    \label{fig:framework}
\end{figure}

\section{Method}

In this section, we introduce the proposed \ours~ that dynamically constructs temporal heterogeneous graphs with clinical calculators and adopts novel GNNs to predict sepsis risks. 

\subsection{Notation and Problem Statement}  
We aim to predict the risk of sepsis with the observed clinical variables.  
We consider the following setup. 
A patient has a sequence of clinical variables (\textit{e.g.}, lab test and vital sign data) with timestamps.  
Let $X \in R^{T \times k}$ denote the observations of variables, where $T$ denotes the number of collections of observations and $k$ denotes the number of unique clinical variables. 
$Y \in \{0, 1\}^T$ denotes the binary ground truth of whether the patient will progress to sepsis in the coming hours. 
We represent patients' data as temporal heterogeneous graphs.
Given a loss function $\mathcal{L}$ and a distribution over pairs ($X$, Y), the goal is to learn the prediction function $f$ by minimizing the expected loss: $f^* = \arg \min_f E [\mathcal{L}(f(X), Y)]$.

\subsection{Static Temporal Graph Construction}

We first construct a static temporal graph to represent the observed variables with timestamps in each patient's EHRs.

\subsubsection{Nodes} 

The temporal graph contains four kinds of nodes. The first is collection nodes which represent a collection of data observed in the same timestamps. 
The second kind of node is the observed clinical variables with the attributes of the observed values. The third kind of node is the organ nodes that describe the organ function statuses. The fourth kind of node is the calculator node that will be added in \autoref{sec:dtgc}. 
 
\subsubsection{Edges}  

We define three kinds of edges between the nodes.
The first is directed edges between successive nodes of the same kind of variable (\textit{e.g.}, black and yellow arrows in \autoref{fig:graph_construction}(C)). The time gaps between two observations are treated as edge attributes.
The second kind of edge is the undirected edge between the nodes with the same timestamps.
The third kind of edge is the directed relation to describe the interactions between clinical events.
Patients might have received treatments (\textit{e.g.}, vasopressor used to avoid low MAP) before the identification of sepsis. Such interaction is important for patients' EHR modeling and incorporated into the graph  (\textit{e.g.}, purple arrows in \autoref{fig:graph_construction}(C)).

\begin{algorithm}[t]
\caption{Temporal Heterogeneous Message Passing} 
\label{alg:thmp}
\leftline{\textbf{Input}:  temporal heterogeneous graph $G$;}
\leftline{\textbf{Output}: collection node feature $\mathbf{h}^L_t$ and organ node feature $\mathbf{h}^{L,i}_t$;}
\begin{algorithmic}[1] 
\STATE Obtain node embedding $\mathbf{h}_v$ and edge embedding $\mathbf{h}_r$ for each node $v$ and edge $r$  with \autoref{eq:node-embedding} and \autoref{eq:edge-embedding};
\FOR{$l \leftarrow 1$ \TO $L$}
\STATE Calculate attention result $\text{Att}^{i}(v_e, v_t)$ with \autoref{eq:attention};
\STATE Concatenate multiple head attention $\mathbf{h}^l_v$ with \autoref{eq:multi-head};
\ENDFOR
\STATE Obtain collection node feature $\mathbf{h}^L_t$ and organ node feature $\mathbf{h}^{L,i}_t$ with \autoref{eq:node_organ_feature};
\STATE Return $\mathbf{h}^L_t$ and $\mathbf{h}^{L,i}_t$.
\end{algorithmic}
\end{algorithm}

\subsection{Temporal Heterogeneous Message Passing}
\label{sec:thmp}

We use a graph encoder with temporal heterogeneous message passing to extract the features from the temporal graphs.  
Algorithm~\autoref{alg:thmp} describes the inference process of the module.

\subsubsection{Clinical Embedding}
We first map all the heterogeneous nodes and edges of the temporal graphs into a same embedding space.





\noindent
\textbf{Node Embedding.} An embedding layer is used to map each node $v$ into a fixed-sized vector $e^v \in R^d$. 
We also embed the observed values as vectors for the nodes with float-value attributes. 
Following~\cite{yin2020identifying}, we adopt value embedding to map the observed value as vector $e^{v'}$ and use time embedding to map the time gap $\delta$ as $e^{\delta}$. 
The concatenation of the node embedding and the value embedding is sent to a linear mapping layer to generate $\mathbf{h}_v \in R^d$,
containing the information of node type, variable name, and observed value:
\begin{equation}
\label{eq:node-embedding}
\mathbf{h}_v = \text{L}([e^v; e^{v'}]), 
\end{equation} 
where $\text{L}(\cdot)$
denotes linear mapping functions, with each instance representing a distinct mapping. $[\cdot; \cdot]$ is a concatenation operation. The details of the embedding can be found in \autoref{sec:embedding}.

\noindent
\textbf{Edge Embedding.} Similarly, we use an embedding layer to map each edge $r$ into a fixed-size vector $e^r \in R^d$. 
For the directed edge $r$ with elapsed time, we combine the edge embedding $e^r$ with the time embedding $e^\delta$ to generate $\mathbf{h}_r \in R^d$:
\begin{equation}
\label{eq:edge-embedding}
\mathbf{h}_r = \text{L}([e^r; e^\delta])
\end{equation} 

\subsubsection{Heterogeneous Message Passing} 
Given the temporal graph and embeddings, we leverage a temporal-aware message-passing mechanism for heterogeneous graphs to effectively gather temporal information, clinical event interaction, and historical observations.

We represent the features of clinical nodes in the $(l)$-th layer of the network as $\mathbf{h}^{(l)}_{v}\in \mathbb{R}^d$. They also serve as the input for the subsequent $(l+1)$-th layer. 

The entire message-passing process can be formalized into two stages: aggregation and updating. The first step is to aggregate the information of neighboring nodes. Specifically, we use an attention mechanism to weigh and integrate the features of neighboring nodes and concatenate the output from multi-head attention to obtain the final message. Taking as an example the process of propagating the features of neighbor nodes $v_e \in N(v_t)$ to the collection node $v_t$, the $i$-th head attention is as follows: 
\begin{align}
&
   \mathbf{q}_v = \text{L}\left(\mathbf{h}^{(l)}_{v_t}\right), 
   \quad 
   \mathbf{W}_r = \text{L}\left(\mathbf{h}_r\right), 
   \quad 
   \mathbf{k}_{v_e} = \text{L}\left(\mathbf{h}^{(l)}_{v_e}\right),
   \notag\\
&\qquad \qquad \text{Att}^{i}(v_e, v_t)  =   \frac{\mathbf{q}_v \mathbf{W}_r \mathbf{k}_e^\top}{\sqrt{d}}, \label{eq:attention}
\end{align}
where $v_e$ represents the clinical node such as lab test, vital sign, and procedure contained in $t$-th collection as well as previous collection node $v_{t-1}$. $r$ denotes the edge between nodes $v_t$ and $v_e$.
The target node for message propagation is $t$-th collection node $v_t$.
After obtaining the attention scores for different neighboring nodes, we combine them with the mapped neighbor features to complete the entire aggregation process.  We formalize it as follows:
\begin{equation}
   \widetilde{\mathbf{h}}^{i}_v = \underset{\forall v_e  \in N(v_t)}{\text{Softmax}}\left(\text{Att}^{i}(v_e, v_t)\right) \cdot \text{L}(\textbf{h}^{(l)}_{v_e}),
\end{equation}
where $N(v_t)$ denotes neighbors of $v_t$, and $ \widetilde{\mathbf{h}}^{i}_v$ represents the message passed to the collection node  by the $i$-th head attention. Following the acquisition of aggregated information, the next stage is to combine this aggregated information with the target node's ego information to update the representation of the target node:
\begin{align}
&\textbf{h}^{(l+1),i}_{v} = \gamma\cdot\text {L }\left(\text{ReLU}\left(\widetilde{\mathbf{h}}^{i}_v\right)\right)+(1-\gamma)\cdot\textbf{h}^{(l)}_{v}, \notag\\
&\textbf{h}^{(l+1)}_{v} =\underset{i\in[1,h]}{\|} \textbf{h}^{(l+1),i}_{v} \label{eq:multi-head}
\end{align} 
where $\underset{i\in[1,h]}{\|}$ denotes concatenating the outputs of multiple heads, $\gamma$ is learnable coefficient for the skip connection. 
Leveraging heterogeneous message passing, we integrate the features of medical events in the collection, along with the graphical structure and historical collection information, into the current collection node, thereby updating the patient's representation.

\subsubsection{Organ-Specific Node and Collection Node Representation}  

After $L$ layers of temporal heterogeneous message passing, we obtain the collection node features $\mathbf{h}^L_{v_{t,c}}$  and six organ node features $\mathbf{h}^L_{v_{t,o,i}} (1\le i \le 6)$. We use linear functions to generate the node representation $\mathbf{h}^{L}_t, \mathbf{h}^{L,i}_t \in R^d$ for further dynamic graph construction:
\begin{equation}
\label{eq:node_organ_feature}
    \mathbf{h}^{L,i}_t = \text{L}(\text{ReLU}(\mathbf{h}^L_{v_{t,o,i}})), 
    \qquad
    \mathbf{h}^{L}_t = \text{L}(\text{ReLU}(\mathbf{h}^L_{v_{t,c}}))
\end{equation}

\subsection{Dynamic Temporal Graph Construction}
\label{sec:dtgc}
As \autoref{fig:framework}(A) shows, 
we add the accurately estimated calculators to temporal graphs. We estimate organ-specific calculators (\textit{e.g.}, DIC~\cite{DIC}) and multi-organ calculators (\textit{e.g.}, SOFA~\cite{vincent1996sofa}) with the same structure. In this subsection, we use multi-organ calculator generation with $\mathbf{h}^{L}_t$ as an example.

\subsubsection{Calculator Score Estimation}

We estimate the values of calculators $e^c_t \in R^c$ for all the $c$ calculators:
\begin{equation}
\label{eq:calculator_estimation}
    e^c_{t} = \text{L}(\mathbf{h}^L_{t})
\end{equation} 

When all the component variables are observed, we have the ground truth for the calculators and use Mean Square Error (MSE) loss to train the calculator estimation module:
\begin{equation}
\label{eq:loss_mse}
    \mathcal{L}_e = \frac{1}{T}\sum_{t=1}^T \frac{1}{\sum_i M_{t,i}}\sum_{i=1}^c (e^c_{t,i} - \hat{e}^c_{t,i})^2 M_{t,i},
\end{equation} 
where $\hat{e}^c_{t,i}$ denotes the ground truth of the clinical calculators. $M_{t,i}$ denotes an indicator variable. $M_{t,i}$ is 1 if the ground truth $\hat{e}^c_{t,i}$ is available, and 0 else.
When the ground truth is not available, the module is jointly trained with the sepsis prediction framework.

\subsubsection{Dynamic Node Constrution}

When missing rates are high, and the generated calculators' scores are inaccurate, consideration of such calculators could introduce additional bias and even mislead clinicians when providing clinical decision-making support. We use a confidence score $p^c_t \in R^c$ to filter out the missing calculators with low confidence (\textit{i.e.}, $p^c_t<0.5$). 
\begin{equation}
    \label{eq:calculator_confidence}
    p^c_t = Sigmoid(\text{L}(\mathbf{h}^L_t))
\end{equation}

We use the following objective to train the dynamic node construction module:
\begin{equation}
\label{eq:construction_loss}
    \mathcal{L}_d = \frac{1}{T}\sum_{t=1}^T \frac{1}{\sum_i M_{t,i}} \sum_{i=1}^c[- y^c_{t,i} \log p^c_{t,i} - (1 - y^c_{t,i}) \log (1- p^c_{t,i})] M_{t,i},
\end{equation}
where $y^c_{t,i} = I[(e^c_{t,i} - \hat{e}^c_{t,i})^2 < 0.01]$ and $I[\cdot]$ is an indicator function that returns 1 if the statement is true; otherwise, 0.

\subsubsection{New Edge Generation}

To incorporate the clinical calculator mechanism, the model also automatically generates the edges between the generated nodes and their component variables.



\begin{algorithm}[t]
\caption{SepsisCalc} 
\label{alg:SepsisCalc}
\leftline{\textbf{Input}: static temporal graph $G$, calculator ground truth $\hat{e}_t$,}
\leftline{\qquad \quad
outcome $y_t$,  learning rate $lr$;}

\begin{algorithmic}[1] 
\REPEAT
\STATE \textbf{\textit{\# Dynamic temporal graph construction}}
\STATE Obtain collection node feature $\mathbf{h}^L_t$ and organ node feature $\mathbf{h}^{L,i}_t$ with  \autoref{alg:thmp} and temporal graph $G$;
\STATE Estimate clinical calculators $e^c_t$ with \autoref{eq:calculator_estimation};
\STATE Compute  calculator estimation loss $\mathcal{L}_e$ with \autoref{eq:loss_mse};
\STATE Estimate calculator confidence $p^c_t$ with \autoref{eq:calculator_confidence};
\STATE Compute calculator confidence loss $\mathcal{L}_d$ with \autoref{eq:construction_loss};
\STATE Obtain dynamic temporal graph $G_d$ by adding calculators with high confidence to temporal graph $G$;
\STATE \textbf{\textit{\# Sepsis risk and organ dysfunction prediction}}
\STATE Obtain collection node feature $\mathbf{h}^D_t$ and organ node feature $\mathbf{h}^{D,i}_t$ with \autoref{alg:thmp} and dynamic graph $G_d$;
\STATE Estimate the sepsis risk $p_t$ and organ dysfunction risk $p^o_t$ with \autoref{eq:risk_estimation};
\STATE Compute prediction loss $\mathcal{L}_c$ and $\mathcal{L}_o$ with \autoref{eq:bceloss};
\STATE Update parameters by minimizing the loss $\mathcal{L}$ in \autoref{eq:sum_loss}; 
\UNTIL{Convergence.}
\end{algorithmic} 
\end{algorithm}

\subsection{Sepsis and Organ Dysfunction Prediction}
After dynamically constructing the temporal graph, we re-extract the features of the new graphs with the same temporal heterogeneous message passing module as in \autoref{sec:thmp}.
We use $\mathbf{h}^D_t$ and $\mathbf{h}^{D,i}_t$  to denote the extracted features for the collection node and $i$-th organ node at time $t$ from the dynamic temporal graph and continue to predict the clinical risk as shown in \autoref{fig:framework}(B).



\subsubsection{Risk Prediction} 
 
We use a linear layer followed with a Sigmoid layer to generate the sepsis risk $p_t \in R$ and organ dysfunction risk $p_t^{o,i} \in R$ ($i=1,2,...,6$): 
\begin{align}
\label{eq:risk_estimation}
    & p_t = Sigmoid( \text{L}(\mathbf{h}^D_t)), \\
    & p^{o,i}_t = Sigmoid( \text{L}(\mathbf{h}^{D,i}_t)),    \notag
\end{align} 

\subsubsection{Objective Prediction}  

We use binary-cross-entropy loss to train the framework:
\begin{align}
\label{eq:bceloss}
    & \mathcal{L}_c = \frac{1}{T}\sum_{t=1}^T  -y_t\log (p_t) - (1-y_t) \log p_t, \\ 
    & \mathcal{L}_o = \frac{1}{T}\sum_{t=1}^T   \frac{1}{6}\sum_{i=1}^6 -y^{o,i}_t\log (p^{o,i}_t) - (1-y^{o,i}_t) \log (1 - p^{o,i}_t) \notag
\end{align} 

The whole framework is trained with a weighted loss:
\begin{equation}
\label{eq:sum_loss}
    \mathcal{L} = \mathcal{L}_c + \alpha_o \mathcal{L}_o + \alpha_e \mathcal{L}_e + \alpha_d \mathcal{L}_d,
\end{equation}
where $\alpha_o, \alpha_e, \alpha_d > 0$ are hyper-parameters. 
Algorithm \autoref{alg:SepsisCalc} describes the training process of the framework.






\begin{figure}
    \centering
    \includegraphics[width=\linewidth]{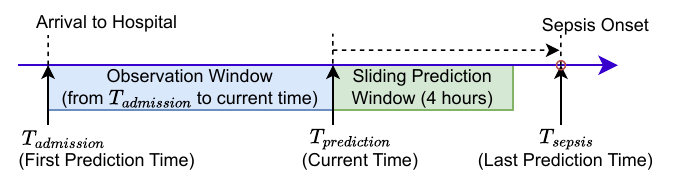}
    \caption{Setting of sepsis onset prediction.
    }
    \label{fig:setting}
\end{figure}

\section{Experiment}

To demonstrate the effectiveness of the proposed \ours, we conducted extensive experiments on multiple real-world datasets.

\vspace{-2mm}
\subsection{Datasets and Setup}
\textbf{Datasets.}
We validated our model on two publicly available datasets 
(MIMIC-III\footnote{\url{https://mimic.physionet.org/}} and AmsterdamUMCdb\footnote{\url{https://amsterdammedicaldatascience.nl}}) and one proprietary dataset extracted from OSUWMC\footnote{https://wexnermedical.osu.edu/}.
We first extracted all the sepsis patients with suspected infection~\cite{sepsis3} in the datasets. 
Patients meeting sepsis-3 criteria~\cite{sepsis3} are defined as case patients, while the others with only suspected infection are treated as control patients. 
Following \cite{aiclinician,yin2020identifying}, we extracted 30 vital signs and lab tests for sepsis patient status modeling. 
The statistics of the three datasets are displayed in \autoref{tab:stati}. The used single-organ and multi-organ calculators are summarized in \autoref{tab:single-organ-cc} and \autoref{tab:multi-organ-cc}.
The used variables and additional details can be found in \autoref{sec:variables}.

\noindent
\textbf{Setup.}
\autoref{fig:setting} displays the setting of the experiments. After the patients arrive at the hospital, we start to predict whether the patients will suffer from sepsis with a sliding 4-hour prediction window. We run the prediction process hourly until the patients have been diagnosed with sepsis or discharged.

\vspace{-2mm}
\subsection{Comparison Methods}

To validate the performance of the proposed \ours~ for sepsis prediction, we implemented various models, including clinical calculator-based methods (i.e., \textbf{NEWS}~\cite{smith2013ability}, \textbf{MEWS}~\cite{subbe2001validation}, \textbf{qSOFA}~\cite{sepsis3}, \textbf{SIRS}~\cite{bone1992definitions}), 
RNN-based methods (\textbf{GRU}~\cite{gru}, \textbf{LSTM}~\cite{lstm}, \textbf{DFSP}~\cite{duan2023early}),
attention-based methods (\textbf{RETAIN}~\cite{retain}, \textbf{Dipole}~\cite{dipole}), graph-based methods (\textbf{GTN}~\cite{chowdhury2023predicting}, \textbf{RGNN}~\cite{liu2020hybrid}).
The details of the comparison methods can be found in \autoref{sec:model_comparison}.

We also implemented various versions of the proposed model.   
    \textbf{\ours} is the main version.
    \textbf{\ours$^{\textbf{\textit{-i}}}$} is the simplest version that uses the same graph construction method as \cite{chowdhury2023predicting,liu2020hybrid} (see \autoref{fig:graph_construction}(B)), without any graph interaction and calculators. 
    \textbf{\ours$^{\textbf{\textit{imp}}}$} uses an imputation method~\cite{3dmice} to replace the dynamic graph construction module. 
    \textbf{\ours$^{\textbf{\textit{-d}}}$} removes  the dynamic graph construction module. \textbf{\ours$^{\textbf{\textit{-o}}}$} removes the organ dysfunction prediction module. 

\vspace{-2mm}
\subsection{Implement Details}

We implement our proposed model with Python 3.8.10 and PyTorch 1.12.1\footnote{\url{https://pytorch.org/}}. For training models, we use Adam optimizer with a mini-batch of 64 patients. The multi-modal data are projected into a $512$-d space. We randomly divide the patients in each dataset into 10 sets. All the experiment results are averaged from 10-fold cross-validation, in which 7 sets are used for training, 1 set for validation, and 2 sets for testing. The validation sets are used to determine the best values of parameters in the training iterations.

For the sepsis prediction tasks, we use Area Under the Receiver Operating Characteristic Curve (AUC), F1 and Recall for evaluation metrics at the collection level (with each collection treated as a separate sample).
For the calculator estimation tasks, following~\cite{3dmice,detroit}, we measure the models' performance with normalized Root Mean Square Error (nRMSE). 
The code and more implementation details can be found in \autoref{sec:implementaion} and GitHub\footnote{\label{github}\url{https://github.com/yinchangchang/SepsisCalc}}.

\begin{table} [!t]
\centering
\caption{Statistics of MIMIC-III, AmsterdamUMCdb, and OSUWMC datasets.} 
\setlength{\tabcolsep}{3pt}
\label{tab:stati} 
\begin{tabular}{cccc}
\toprule
 & MIMIC & AmsterdamUMCdb  & OSUWMC\\
 \midrule
 \#. of patients & 21,686 & 6,560 & 85,181\\
 \#. of male & 11,862 & 3,412& 41,710\\
 \#. of female & 9,824 & 3,148& 43,471\\
 Age (mean $\pm$ std)~~~ & 60.7 $\pm$ 11.6 & 62.1 $\pm$ 12.3 & 59.3 $\pm$ 16.1 \\
 Missing rate & 65\% & 68\% & 75\%\\
 Sepsis rate & 32\% & 35\% & 29\%\\
 \bottomrule
\end{tabular} 
\end{table} 

\section{Results}

\label{sec:result}
We now report the performance of the proposed model in the three datasets.
We focus on answering the following research questions using our experimental results:
\begin{itemize}
    \item \textbf{Q1: Why must we incorporate the clinical calculator scores?}
    \item \textbf{Q2: Are the estimated clinical calculator scores effective?} 
    \item \textbf{Q3: How do estimated calculator scores improve early sepsis prediction system?}  
\end{itemize} 

\vspace{-1mm}
\subsection{Q1: Why must we incorporate the clinical calculator scores?}
 
\subsubsection{Wide Adoption of Clinical Calculators}
Clinical calculators have emerged as indispensable tools within healthcare settings, providing clinicians with evidence-based risk assessments essential for accurate diagnosis and prognostic evaluation \cite{jin2024agentmd,dziadzko2016clinical,green2019medical}.
 In the context of sepsis, numerous calculators, such as SOFA \cite{sepsis3,vincent1996sofa}, qSOFA \cite{paoli2018epidemiology,dorsett2017qsofa}, MEWS \cite{subbe2001validation}, NEWS \cite{smith2013ability}, and SIRS \cite{bone1992definitions}, are extensively studied in existing sepsis-related literature~\cite{guan2021survival,giamarellos2017validation,marik2017sirs,qiu2023sirs}, and widely employed as early warning tools in real-world hospital EHR systems for both ICU and hospital wards \cite{sepsis3,bhattacharjee2017identifying}. 
Integrating clinical calculators into early sepsis prediction models can align them more closely with clinicians' workflows and enhance comprehensibility.

\subsubsection{Improvement from Calculators on Sepsis Prediction Performance}
Compared to raw clinical variables, clinical calculators can summarize the patients' health states at a high level (\textit{e.g.}, in single or multiple organ dysfunction levels).
We first design experiments to demonstrate the effectiveness of clinical calculators on early sepsis prediction with four settings:
\begin{itemize}
    \item \textbf{Full Observation without Clinical Calculator (FO)}: All the component variables of calculators are available while the clinical calculators are not used.
    \item \textbf{Full Observation with Clinical Calculator (FOCC)}: All the component variables of calculators are available and the clinical calculators are also included when conducting sepsis prediction.
    \item \textbf{Missing Observation without Clinical Calculator (MO)}: Only partial component variables of clinical calculators are observed and the calculators are not incorporated.
    \item \textbf{Missing Observation with Clinical Calculator (MOCC)}: Partial component variables of clinical calculators are observed and we still use the variables to compute the calculators.
\end{itemize}   

\begin{figure}
    \centering
    \includegraphics[width=0.48\linewidth]{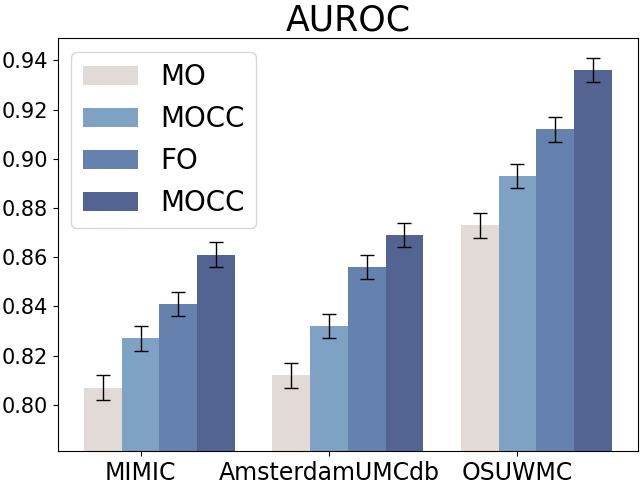}
    \includegraphics[width=0.48\linewidth]{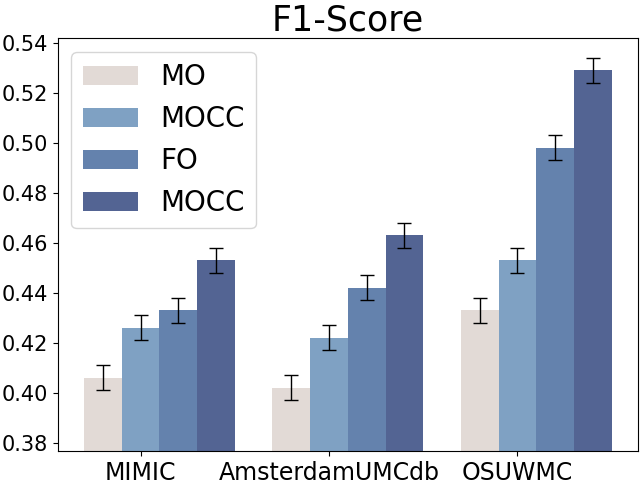}
    \caption{Sepsis risk prediction performance in both full and missing observation settings.}
    \label{fig:full_missing_setting}
\end{figure}

\autoref{fig:full_missing_setting} displays the results for sepsis prediction in the four settings. The results show that in both full observation and missing observation settings, clinical calculators (\textit{i.e.}, FOCC and MOCC) can consistently improve sepsis prediction performance, which shows the effectiveness of such domain knowledge in clinical tasks.

    

\begin{figure}
    \centering
    \includegraphics[width=0.48\linewidth]{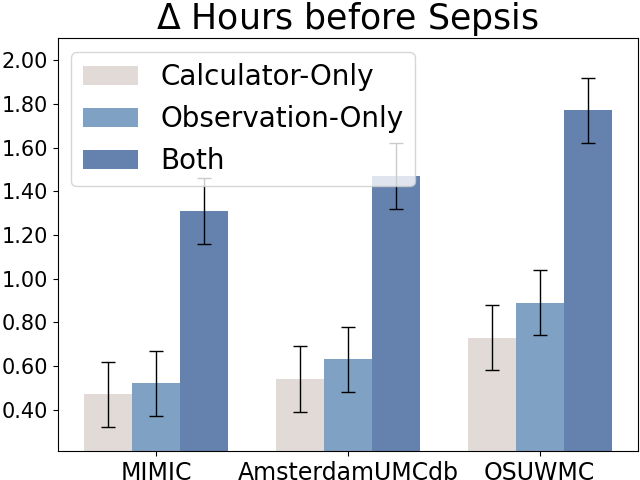}
    \includegraphics[width=0.48\linewidth]{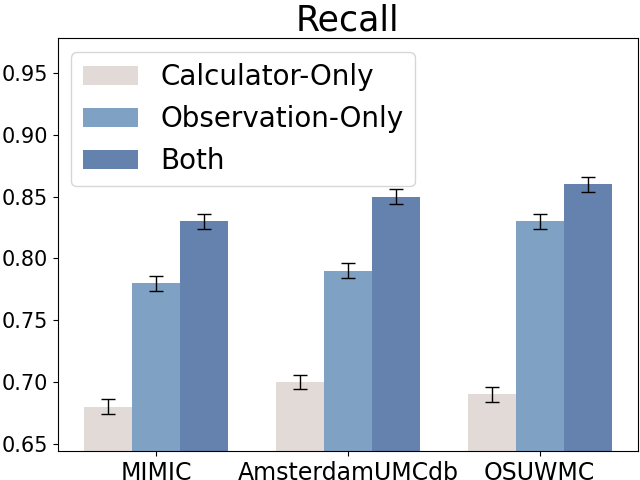}
    \vspace{-3mm}
    \caption{Average alert time before sepsis and recall.}
    \vspace{-3mm}
    \label{fig:time_before_sepsis}
\end{figure}

\subsubsection{Early Sepsis Alerts with Clinical Calculators.}  
Due to the fast-development characteristics of sepsis, delayed identification and treatment will significantly reduce the patients' survival rates~\cite{liu2017timing}. It is critical to early identify the patients with sepsis risks.
We also reported the average sepsis alert time before sepsis and recall at the patient level (with each patient treated as a separate sample for evaluation), as shown in \autoref{fig:time_before_sepsis}. The sepsis prediction models that incorporate both raw observations and calculators can predict sepsis approximately 1 hour earlier and achieve higher recall compared to models using only raw observational data, further demonstrating the credibility and effectiveness of the calculators.

\subsection{Q2: Are the estimated clinical calculator scores effective?}

\subsubsection{Misssing Rate of Clinical Calculators} 
Directly incorporating the clinical calculators might not be applicable for two reasons: (i) Lots of risk calculators (\textit{e.g.}, SOFA) aggregate the values of clinical variables in a specific time span and thus are not immediately available, which limits their usage in timely sepsis prediction and detection.
(ii) Most risk calculators usually combine the information from multiple variables. In real-world settings, the variables might have high missing rates and not always be available (especially for blood lab tests).    
\autoref{tab:all-variables} in Appendix displays the missing rates of the important lab tests and calculators related to sepsis.  

\subsubsection{Effectiveness of Clinical Calculator Estimation}
To address the problem, we propose to estimate the calculators. In this subsection, we evaluated the performance of clinical calculator estimation. 

\noindent
\textbf{Full Observation Setting.}
When all the component variables are observed, the ground truths of the clinical calculators are available, we used nRMSE to evaluate the calculator estimation performance. \autoref{tab:nrmse_gt} in Appendix shows that the nRMSE between the ground truths and the estimated calculators is close to 0, demonstrating that our model can accurately learn the computation mechanisms of clinical calculators.

\begin{figure}
    \centering
    \includegraphics[width=0.48\linewidth]{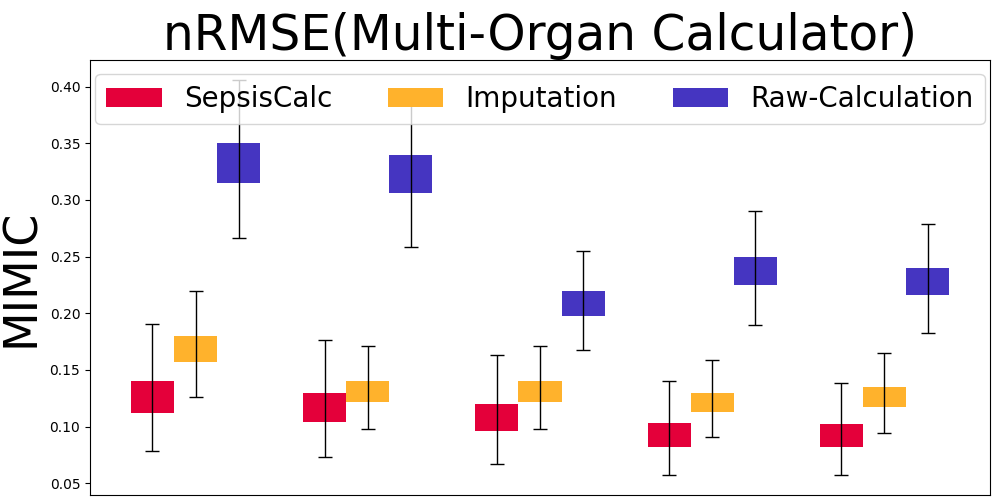}
    \includegraphics[width=0.48\linewidth]{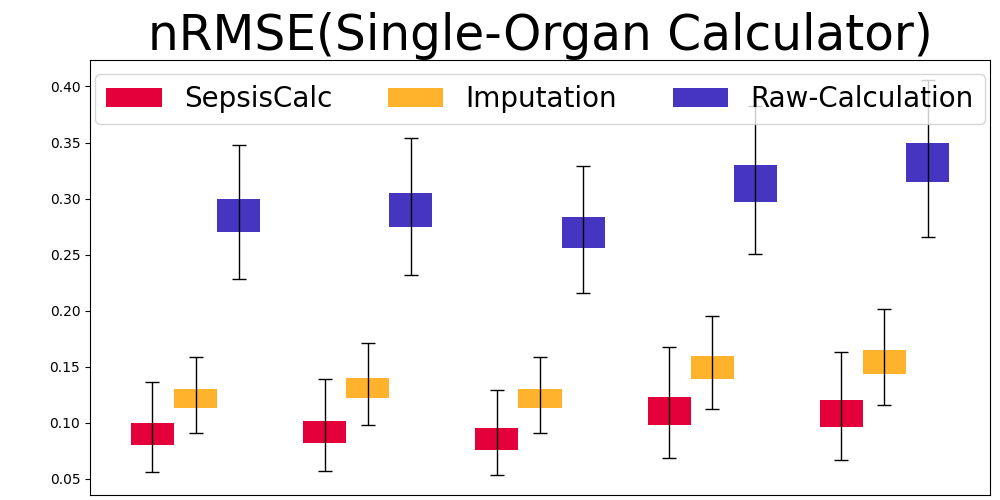}
    \includegraphics[width=0.48\linewidth]{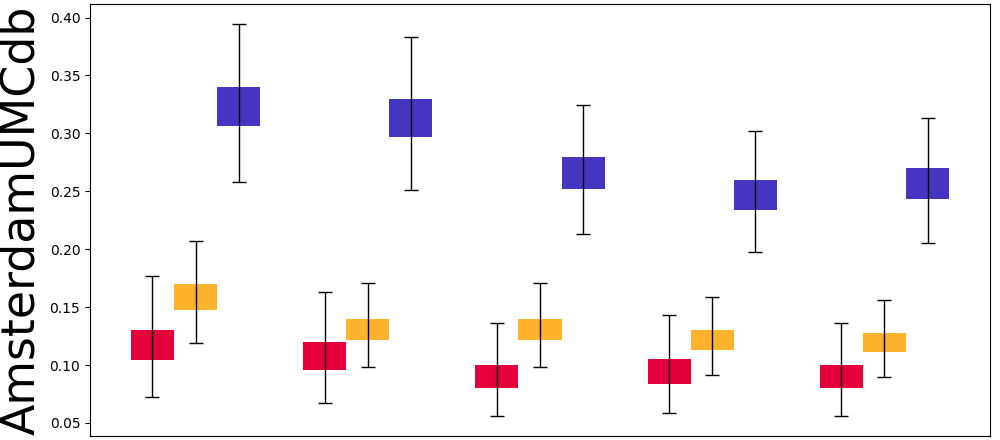}
    \includegraphics[width=0.48\linewidth]{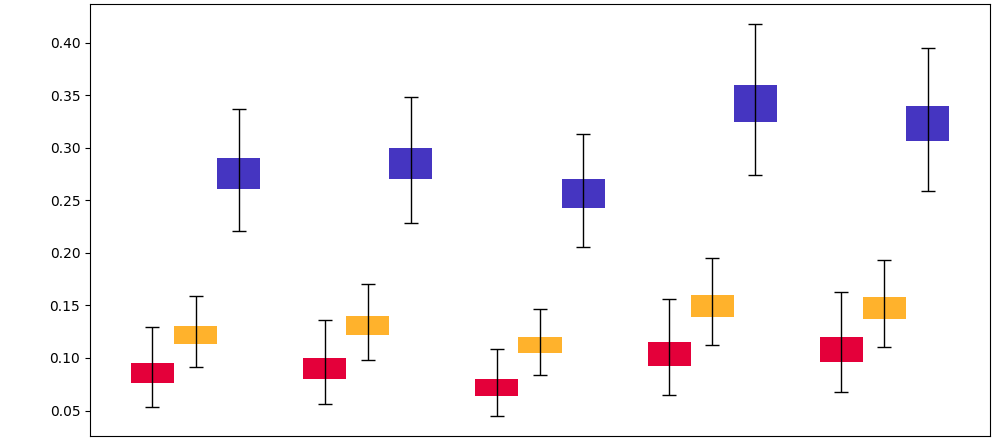}
    \includegraphics[width=0.48\linewidth]{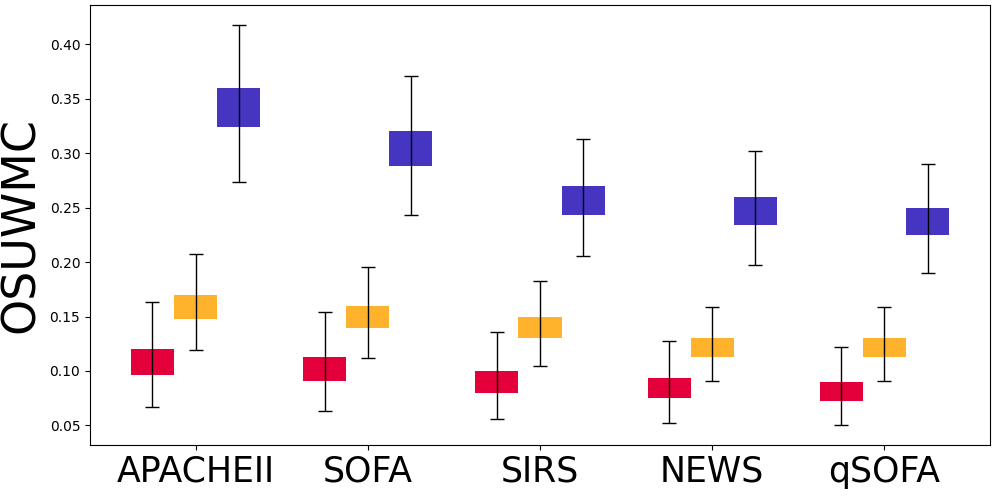}
    \includegraphics[width=0.48\linewidth]{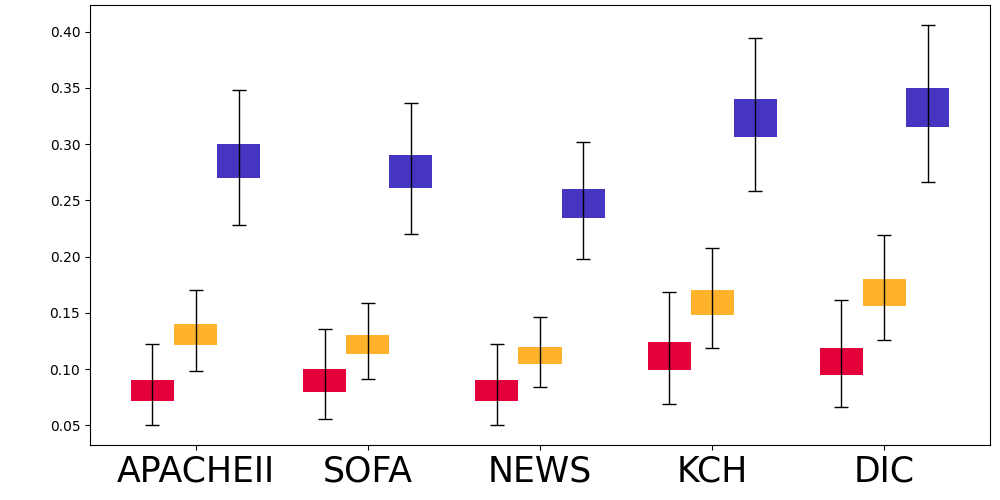}
    \caption{nRMSE of clinical calculator estimation (mask observation setting). All the component variables of the calculators are observed and the ground truths of calculators are available. We randomly mask 70\% component variables. Raw-calculation means the original clinical methods that use the latest observed variables to compute the calculators. }
    \label{fig:nrmse_mask}
\end{figure}

\begin{figure}
    \centering
    \includegraphics[width=0.48\linewidth]{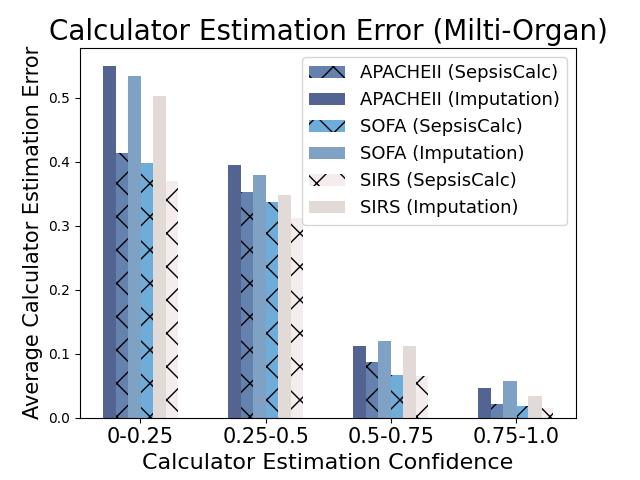}
    \includegraphics[width=0.48\linewidth]{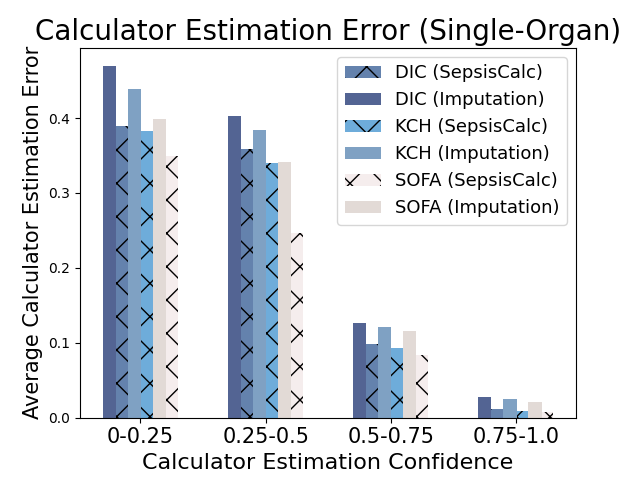}
    \caption{Calculator estimation error over confidence levels. }
    \label{fig:cc_error}
\end{figure}

\begin{table*}[]
    \centering
    \caption{Sepsis prediction results. }
    \begin{tabular}{c|ccc|ccc|ccc}
    \toprule
        & \multicolumn{3}{c|}{MIMIC-III}& \multicolumn{3}{c|}{AmsterdamUMCdb} & \multicolumn{3}{c}{OSUWMC} \\
        Method & AUC & F1 & Recall & AUC & F1 & Recall & AUC & F1 & Recall \\
        
    \midrule

    NEWS   & 0.722$_{\pm0.012}$ & 0.366$_{\pm0.013}$ & 0.620$_{\pm0.012}$ & 0.731$_{\pm0.013}$ & 0.370$_{\pm0.013}$ & 0.627$_{\pm0.013}$ & 0.765$_{\pm0.012}$ & 0.387$_{\pm0.013}$ & 0.656$_{\pm0.012}$\\
    MEWS   & 0.726$_{\pm0.013}$ & 0.372$_{\pm0.013}$ & 0.624$_{\pm0.012}$ & 0.735$_{\pm0.012}$ & 0.376$_{\pm0.012}$ & 0.631$_{\pm0.013}$ & 0.769$_{\pm0.013}$ & 0.393$_{\pm0.012}$ & 0.660$_{\pm0.013}$\\
    qSOFA  & 0.729$_{\pm0.011}$ & 0.374$_{\pm0.011}$ & 0.631$_{\pm0.011}$ & 0.738$_{\pm0.012}$ & 0.379$_{\pm0.012}$ & 0.639$_{\pm0.012}$ & 0.772$_{\pm0.011}$ & 0.396$_{\pm0.011}$ & 0.668$_{\pm0.011}$\\
    SIRS   & 0.733$_{\pm0.011}$ & 0.376$_{\pm0.012}$ & 0.642$_{\pm0.011}$ & 0.742$_{\pm0.012}$ & 0.381$_{\pm0.012}$ & 0.650$_{\pm0.011}$ & 0.776$_{\pm0.012}$ & 0.398$_{\pm0.012}$ & 0.680$_{\pm0.012}$\\

    \midrule
    GRU    & 0.801$_{\pm0.012}$ & 0.397$_{\pm0.012}$ & 0.696$_{\pm0.013}$ & 0.807$_{\pm0.012}$ & 0.400$_{\pm0.012}$ & 0.701$_{\pm0.012}$ & 0.872$_{\pm0.012}$ & 0.432$_{\pm0.012}$ & 0.758$_{\pm0.012}$\\
    LSTM   & 0.807$_{\pm0.013}$ & 0.408$_{\pm0.012}$ & 0.698$_{\pm0.012}$ & 0.813$_{\pm0.012}$ & 0.411$_{\pm0.012}$ & 0.703$_{\pm0.012}$ & 0.879$_{\pm0.013}$ & 0.444$_{\pm0.012}$ & 0.760$_{\pm0.012}$\\
    RETAIN & 0.814$_{\pm0.013}$ & 0.418$_{\pm0.013}$ & 0.710$_{\pm0.014}$ & 0.820$_{\pm0.013}$ & 0.421$_{\pm0.013}$ & 0.715$_{\pm0.014}$ & 0.886$_{\pm0.014}$ & 0.455$_{\pm0.013}$ & 0.773$_{\pm0.013}$\\
    Dipole & 0.817$_{\pm0.014}$ & 0.423$_{\pm0.013}$ & 0.703$_{\pm0.013}$ & 0.823$_{\pm0.014}$ & 0.426$_{\pm0.014}$ & 0.709$_{\pm0.014}$ & 0.889$_{\pm0.013}$ & 0.461$_{\pm0.013}$ & 0.766$_{\pm0.014}$\\
    DFSP  & 0.822$_{\pm0.011}$ & 0.424$_{\pm0.012}$ & 0.696$_{\pm0.011}$ & 0.828$_{\pm0.011}$ & 0.425$_{\pm0.011}$ & 0.702$_{\pm0.011}$ & 0.894$_{\pm0.012}$ & 0.465$_{\pm0.012}$ & 0.758$_{\pm0.012}$\\
    \midrule
    RGNN & 0.819$_{\pm0.013}$ & 0.424$_{\pm0.012}$ & 0.698$_{\pm0.012}$ & 0.825$_{\pm0.013}$ & 0.427$_{\pm0.013}$ & 0.703$_{\pm0.013}$ & 0.892$_{\pm0.013}$ & 0.467$_{\pm0.012}$ & 0.760$_{\pm0.012}$\\
    GTN  & 0.821$_{\pm0.014}$ & 0.423$_{\pm0.014}$ & 0.707$_{\pm0.013}$ & 0.827$_{\pm0.013}$ & 0.426$_{\pm0.014}$ & 0.712$_{\pm0.014}$ & 0.893$_{\pm0.013}$ & 0.465$_{\pm0.014}$ & 0.770$_{\pm0.014}$\\
    \midrule
    \ours  & \textbf{0.839$_{\pm0.011}$} & \textbf{0.438$_{\pm0.012}$} & \textbf{0.729$_{\pm0.011}$} & \textbf{0.848$_{\pm0.012}$} & \textbf{0.442$_{\pm0.011}$} & \textbf{0.735$_{\pm0.012}$} & \textbf{0.918$_{\pm0.012}$} & \textbf{0.479$_{\pm0.011}$} & \textbf{0.791$_{\pm0.012}$}\\
    \bottomrule
    \end{tabular}
    \label{tab:sepsis_prediction_results}
\end{table*} 

\noindent
\textbf{Mask Observation Setting.}
To further validate the calculator estimation performance when component variables are missing, we randomly masked 70\% component variables to keep them consistent with real-world missing rates (see \autoref{tab:all-variables}). \autoref{fig:nrmse_mask} displays the results of our \ours, imputation~\cite{3dmice} and original calculator computation mechanisms (\textit{i.e.}, Raw-Calculation). \ours~ achieved much smaller estimation errors than Raw-Calculation, leading to the improved performance of downstream sepsis prediction tasks.  

When component variables are missing, the ground truths of the clinical calculators are not available. Instead, we use the performance of downstream sepsis prediction tasks to validate the effectiveness of calculator estimation tasks. The detailed results are displayed in \autoref{sec:improment_cc}.

\subsubsection{Effectiveness of Clinical Calculator Generation Confidence}
When the missing rates are relatively high, the estimated clinical calculators might be inaccurate. We propose to use the calculator estimation confidence $p^c_t$ in \autoref{eq:calculator_confidence} to filter the inaccurately estimated calculator nodes (\textit{i.e.}, $p^c_t<0.5$).  
\autoref{fig:cc_error} shows that when the confidence of generation is low, calculator estimation performance suffers from a significant decline. Existing imputation models always give imputation results for the missing values, which might introduce more imputation bias to the prediction models and could be harmful for high-stake clinical applications. Moreover, the error-prone imputation in high-missing-rate settings could further mislead clinicians when providing clinical decision-making support. 
Our dynamic graph construction module only generates the clinical calculators with high confidence, which achieves a good trade-off between introducing more domain-specific knowledge and reducing imputation bias.

\subsection{Q3: How do estimated calculator scores improve early sepsis prediction system?}

\subsubsection{Early Sepsis Prediction} 
\autoref{tab:sepsis_prediction_results} displays the sepsis prediction results. All the deep learning methods outperform the early-warning scores (i.e., NEWS, MEWS, qSOFA, SIRS), which shows the promising potential of state-of-the-art deep learning models in real-world clinical applications.
Although human-designed calculators are effective, deep-learning methods can capture abnormal values and more complicated temporal patterns inside EHRs. 

Compared with attention-based models and graph-based models, the proposed \ours~  achieved the best prediction performance. By considering both observations and the estimated clinical calculators, the proposed \ours~ can model the organ dysfunctions better, which further improves the performance.
The combination of human-designed clinical calculators and end-to-end deep learning methods can not only achieve better performance but also enhance credibility in real-world applications.

\subsubsection{Ablation Study}
\label{sec:improment_cc}

\begin{table}[]
    \centering
    \setlength{\tabcolsep}{1.2pt}
    \caption{Ablation study.}
    \begin{tabular}{l|ccc|ccc|ccc}
    \toprule
        Method & \multicolumn{3}{c|}{MIMIC-III} & \multicolumn{3}{c|}{\small AmsterdamUMCdb} & \multicolumn{3}{c}{OSUWMC}\\ 
        & AUC & F1 & \small{Recall} & AUC &  F1 & \small{Recall} & AUC & F1  & \small{Recall} \\
        \midrule
        \small \ours$^{imp}$ & .825 & .426 & .713 & .835 & .433 & .719 & .908 & .466 & .775 \\
        \small \ours$^{-i}$ & .820 & .422 & .712 & .829 & .427 & .716 & .900 & .463 & .772 \\
        \small \ours$^{-d}$ & .830 & .427 & .715 & .839 & .432 & .715 & .910 & .468 & .775 \\
        \small \ours$^{-o}$ & .831 & .429 & .718 & .841 & .431 & .716 & .911 & .465 & .781 \\
         \small \ours       & \textbf{.839} & \textbf{.438} & \textbf{.729} & \textbf{.848} & \textbf{.442} & \textbf{.735} & \textbf{.918} & \textbf{.479} & \textbf{.791} \\
        \bottomrule
    \end{tabular}
    \label{tab:missing_cal_sepsis_prediction}
\end{table}

To validate the performance improvement from dynamic calculator generation, we conducted an ablation study with various versions of the proposed model.
\autoref{tab:missing_cal_sepsis_prediction} displays the ablation study results.  
Without the clinical event interaction, \ours$^{-i}$ performs worse than other versions, demonstrating the effectiveness of the proposed static graph construction method (\textit{i.e.}, with medical event interaction and successive connection of the same variables).
\ours~ outperforms \ours$^{-d}$ (without dynamic calculator estimation), demonstrating the effectiveness of clinical calculators in sepsis prediction tasks.
\ours~ outperforms \ours$^{imp}$ that use imputation for further calculator computation, demonstrating the effectiveness of the dynamic graph construction. We speculate the reason is that the imputation results might be not accurate during the high-missing-rate settings and introduce more bias to the downstream sepsis prediction tasks, while the proposed \ours~ only includes the accurately estimated calculators with high confidence in the graphs.
By adding the multi-task learning for both organ dysfunction and sepsis risk prediction tasks, \ours~ performs better than \ours$^{-o}$, further showing that the organ dysfunction identification tasks can help the sepsis prediction tasks.









\begin{figure*}
    \centering
    \includegraphics[width=0.85\linewidth]{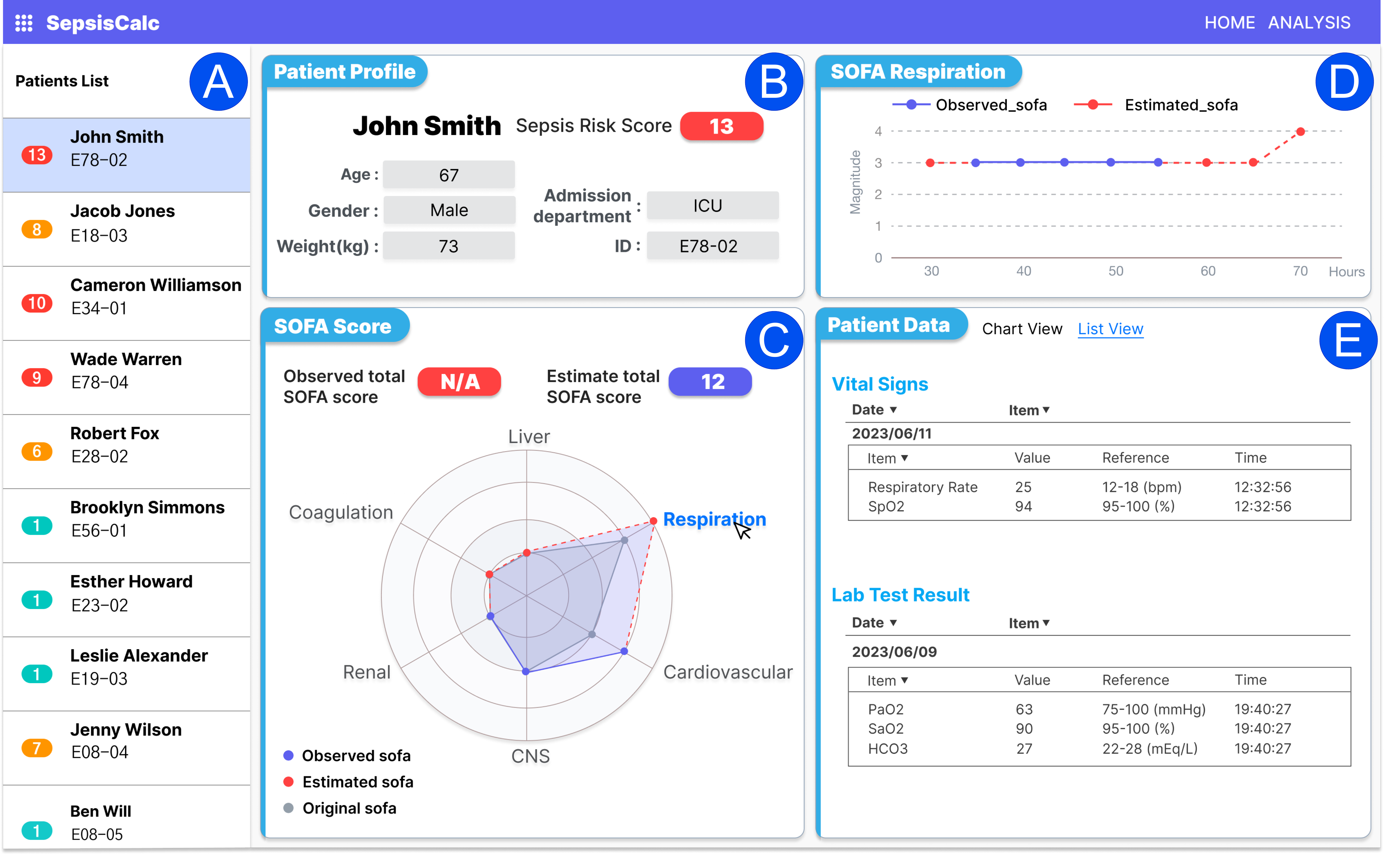}
    \caption{User Interface of \ours~ System. (A) Patient list with sepsis risk score. (B) Demographic information. (C)  Overall SOFA scores. (D) Organ-specific SOFA score. (E) Vital signs and lab test results related to the specific organ.}
    \label{fig:UI}
\end{figure*}

\section{Deployment}

\label{sec:deployment} 
 

Based on the sepsis prediction model, we implemented a system deployed in the Epic EHR Systems\footnote{\label{epic}\url{https://www.epic.com/software/}} at OSUWMC (see \autoref{fig:UI}).  
The system starts to collect patients' data after the patients arrive at hospitals and automatically predicts sepsis risks hourly.

The interactive process with our system is visualized in the provided UI (Figure~\ref{fig:UI}). In this scenario, a clinician is examining high-risk patients. After reviewing the patient list (Figure~\ref{fig:UI}(A)), the clinician selects a patient, prompting the Patient Demographics (Figure~\ref{fig:UI}(B)) section update to display his profile, including age, gender, weight, admission department, and sepsis risk score.

The clinician then focuses on the SOFA Score (Figure~\ref{fig:UI}(C)) to assess the overall organ dysfunctions. Due to the missing variables, the observed total SOFA score is not applicable, and our \ours~ estimates the SOFA score as 12.
The radar chart provides a visual summary of scores across different organs. 
The gray shaded area displays the original SOFA scores at the beginning of the ICU admission, while the blue shaded area represents the observed SOFA (solid blue lines) and estimated SOFA (dashed red lines) scores.

To delve deeper, the clinician clicks the respiration area in the radar chart and reviews how the SOFA score for respiration has evolved in \autoref{fig:UI}(D), and the details of relevant vital signs (\textit{i.e.}, respiratory rate and SpO2), and lab test results (e.g., PaO2, SaO2, HCO3)  in \autoref{fig:UI}(E). 
We also provide a trend view to display vital signs and lab tests so clinicians can review the history of variables and critical changes in \autoref{fig:detail-UI}(B) in Appendix.

The clinician can also examine other organ SOFA scores and their specific details from the radar chart. By clicking on different areas of the radar chart, the clinician can view the scores and details for the central nervous system (CNS), coagulation, renal, liver, and cardiovascular systems. This allows for a comprehensive assessment of each organ's functional status, thereby facilitating more precise clinical decision-making.

Note that we used OSUWMC data for our algorithm illustration.
All patients' names and demographic info in this \autoref{fig:UI} are randomly generated for illustration purposes. 
Ongoing deployment also includes recruiting clinicians for usability evaluation to quantitative and qualitatively measure clinical outcomes and user satisfaction of SepsisLab (OSUWMC IRB\#: 2020H0018).

\vspace{-1mm}
\section{Conclusion}
In this work, we aim to develop transparent and convincing models for the real-world sepsis prediction tasks.
We propose a novel framework \ours, which represents patients' EHR data as dynamic temporal graphs and effectively extracts temporal information, clinical event interactions, and organ dysfunction information from the graphs with a temporal heterogeneous message passing module.
We introduce a dynamic graph construction module to estimate and integrate clinically widely used calculators into sepsis prediction models to help assess organ dysfunctions, aligning well with clinicians' workflows for sepsis identification. Our graph construction method naturally handles the missing values by including only observed variables and high-confidence calculators, thereby avoiding the potential biases introduced by imputation methods that most sepsis prediction models suffer from.
Experiments on three real-world datasets show that \ours~ can not only accurately estimate the calculators to assess the organ dysfunctions (even with missing values), but also outperform state-of-the-art clinical risk prediction methods, demonstrating the effectiveness of \ours. Finally, we design a system to display the identified organ dysfunctions and potential sepsis risks, providing a human-AI interaction tool for deployment and paving the way for actionable clinical decision-making support for early intervention.

\section{ACKNOWLEDGMENTS}
This work was funded in part by the National Science Foundation under award number IIS-2145625, by the National Institutes of Health under award number R01AI188576.
\bibliographystyle{ACM-Reference-Format}
\bibliography{sample-base}

\appendix

\appendix
\section{Appendix}

\subsection{Important Notations}
We summarize the important notations in \autoref{tab:notations}.
\begin{table}[!th]
    \centering
    \caption{Important notations. }
    \vspace{-1mm}
    \begin{tabular}{l|l}
        \toprule
        Notation & Description \\
        \midrule
            
         $X \in R^{T \times k}$ & The original observed EHRs. \\
          $Y \in \{0, 1\}^{T}$&  The ground truth for sepsis prediction.\\
          $T$ & The number of collections of observations.\\
          $t$ & The $t$-th collection. \\
          $k$ & The number of unique variables.\\
          $c$ & The number of unique clinical calculators.\\
          $\delta_{i,j} \in R$ & The time gap between $i$-th and $j$-th collections. \\
          $v$ & A specific node in graph.\\
          $r$ & A specific edge in graph.\\
          $e^v \in R^d$ & The embedding vector for node $v$.\\
          $e^r \in R^d$ & The embedding vector for edge $r$.\\
          $d$ & The dimension of embedding vectors. \\ 
          $\mathbf{h}_v\in R^d$ & The node embedding (with observed value). \\
          $\mathbf{h}_r \in R^d$ & The edge embedding (with time gaps). \\
          $\mathbf{h}^{(l)}_v\in R^d$ & The $l$-th layer feature for node $v$. \\
          $\mathbf{h}^L_t$ & $t$-th collection node's final features.\\
          $\mathbf{h}^{L,i}_t$ & $i$-th organ node's final features ($t$-th collection).\\
          
          $\text{L}(\cdot)$ & The linear mapping function.\\
          $N(v_t)$ & The neighbors of node $v_t$. \\
          $0 <  \gamma < 1$ & The coefficient for the skip connection. \\
          $e^c_t \in R^c$ & The estimated calculators in $t$-th collection.\\ 
          $\hat{e}^c_t \in R^c$ & The ground truth for $e^c_t$.\\
          $M_{t,i}$ & Binary variable indicating the availability of $\hat{e}^c_{t,i}$. \\
          $p^c_t$ & The estimated confidence for $e^c_t$. \\
          $y^c_t$ & The ground truth for $p^c_t$. \\
          $p_t$ & The predicted sepsis probability in $t$-th collection. \\
          $y_t$ & The ground truth of $p_t$. \\ 
          $p^{o,i}_t$ & The organ dysfunction probability for organ $i$. \\
          $y^{o,i}_t$ & The ground truth of $p^{o,i}_t$. \\
          $\mathcal{L}_*$ & Loss functions. \\          
         \bottomrule 
    \end{tabular}
    \vspace{-4mm}
    \label{tab:notations}
\end{table}

\subsection{Clinical Calculators}

\subsubsection{SOFA Score Calculation}

SOFA (Sepsis-related Organ Failure Assessment) score is widely used to describe organ dysfunction for septic patients in real-world clinical settings and is displayed to help clinicians assess the patients' health states in our system~\autoref{fig:UI}. We display the detailed computation method in Table \ref{tab:sofa}.
Each organ's SOFA score ranges from 0 (normal) to 4 (most abnormal). The total SOFA score ranges from 0 (normal) to 24 (most abnormal).
Although SOFA is a multi-organ dysfunction calculator, we can use the component scores to assess specific organ dysfunction when partial variables are missing.

\subsubsection{Multi-Organ Calculators}

In this study, we integrate multiple widely-used and well-validated clinical calculators related to sepsis, including SOFA~\cite{vincent1996sofa}, qSOFA~\cite{sepsis3}, APACHE II~\cite{knaus1985apache},  SIRS~\cite{bone1992definitions}, NEWS~\cite{smith2013ability}, and MEWS~\cite{subbe2001validation}.
The calculators can effectively describe the overall health status of critically ill patients (\textit{e.g.}, sepsis patients) by assessing multiple organ dysfunctions.  \autoref{tab:multi-organ-cc} displays the component variables for various organ function assessments. 

\subsubsection{Single-Organ Calculators}
When the overall calculator is not applicable due to missing values, we can still use the observed variables to compute the organ-specific calculators (\textit{e.g.}, PaO2 and FiO2 for the respiration system in SOFA as shown in \autoref{tab:sofa}). In the six organs in \autoref{tab:multi-organ-cc}, the variables for coagulation and liver systems (\textit{i.e.}, PLT, PT, and Bili) have relatively higher missing rates. We incorporate multiple organ-specific calculators (\textit{e.g.}, DIC~\cite{DIC} for coagulation, KCH~\cite{KCH}, MELD\cite{MELD}, CPS~\cite{CPS} for liver) to assess the corresponding organ status. 
\autoref{tab:single-organ-cc}  presents the single-organ clinical calculators utilized in this work.

Note that the proposed \ours~  can handle various clinical calculators and can be further enhanced with the inclusion of more useful and related calculators.

\begin{table}
    \centering
    \setlength{\tabcolsep}{2pt}
    \caption{Single-organ clinical calculators. PLT: Platelet, Bili: Bilirubin, Enc: Encephalopathy, PT: Prothrombin Time, Fbg: Fibrinogen, DD: D-Dimer, FDPs: Fibrin Degradation Products, KB: Ketone Bodies, LCT: Lactate.}
    \begin{tabular}{l|lll}
    \toprule
         Calculators & Variables & Organ & Range\\
    \midrule 
    AKIN~\cite{AKIN} & Creatinine, Urine output & Renal & 0-3 \\  
    KDIGO~\cite{KDIGO} & Creatinine, Urine output & Renal & 0-3 \\
    \midrule
    KCH~\cite{KCH} & PT, Bili, KB, LCT, Sodium & Liver & 0-20 \\
    MELD~\cite{MELD} & Bili, INR, Creatinine & Liver & 6-40\\
    CPS~\cite{CPS} & Bili, Albumin, PT, Asc, Enc& Liver & 5-15\\
    \midrule
    DIC~\cite{DIC} & PLT, PT, APTT, Fbg, DD, FDPs & Coagulation& 0-12\\
    \bottomrule
    \end{tabular}
    \label{tab:single-organ-cc}
\end{table}

\begin{table*} [!htb]
\centering
\caption{The definition of SOFA score and its components across six organ systems. Each SOFA component score ranges from 0 (normal) to 4 (most abnormal). The total SOFA score ranges from 0 (normal) to 24 (most abnormal).} 
\label{tab:sofa} 
\begin{tabular}{lllll}
\toprule
SOFA score & 1 & 2 & 3 & 4 \\
\midrule
Respiration \\
PaO$_2$/FiO$_2$, mmHg & $<400$ & $<300$ & $<200$ & $<100$ \\
\midrule
Coagulation \\
Platelets $\times 10^3$ /mm$^3$ & $<150$ & $<100$ & $<50$ & $<20$ \\
\midrule
Liver \\
Bilirubin, mg/dl & 1.2 - 1.9 & 2.0 - 5.9 & 6.0 - 11.9 & $>12.0$ \\
($\mu$mol/l)     & (20 - 32) & (33 - 101) & (102 -  204) & ($>204$) \\
\midrule
Cardiovascular \\
Hypotension & MAP $<$ 70 mmHg & Dopamine $\leq$ 5 & Dopamine $>$ 5 & Dopamine $>$ 15 \\
            &                 & or dobutamine (any dose)  & or epinephrine $\leq$ 0.1 & or epinephrine $>$ 0.1 \\
            &                 &                           & or norepinephrine $\leq$ 0.1 & or norepinephrine $>$ 0.1 \\
\midrule
Central nervous system  (CNS) \\
Glasgow Coma Score (GCS) &    13 - 14     &     10 - 12      &    6 - 9       &  $<$6 \\
\midrule
Renal \\
Creatinine, mg/dl  &  1.2 - 1.9 &  2.0 - 3.4 & 3.5-4.9  & $>$ 5.0  \\
($\mu$mol/l) or urine   & (110 - 170) & (171 - 299) &   (300 - 440)   & ($>$ 440) \\
output  & & & or $<$ 500 ml/day & or $<$200 ml/day \\
\bottomrule
\end{tabular} 
\end{table*}

\begin{table*}[]
    \centering
    \caption{Multi-organ clinical calculators. Temp: Temperature, RR: Respiratory Rate, HR: Heart Rate, Bili: Bilirubin, DA: Dopamine, DOB: Dobutamine, EPI: Epinephrine, NE: Norepinephrine, SS: Serum Sodium, SP: Serum Potassium, PLT: Platelets.}
    \begin{tabular}{l|llllll|ll}
    \toprule
         Calculators & Respiration & Coagulation & Liver & Cardiovascular & CNS & Renal & other  & Range\\
    \midrule
    SOFA~\cite{vincent1996sofa} & PaO2, FiO2 & PLT & Bili & MAP, DOB,   & GCS & Creatinine,  &     & 0-24\\
                                &            &           &           & DA, EPI, NE &     & Urine output &      &         \\ 
    \midrule
    qSOFA~\cite{sepsis3}        & RR & & & SBP & GCS & &  & 0-3\\
    \midrule
 SIRS~\cite{bone1992definitions}& RR, PaCO2  &           &           & HR                          &     &              & WBC, Temp  & 0-4\\
    \midrule
    NEWS~\cite{smith2013ability}& RR, SpO2   &           &           & HR, SBP                     & GCS &              & Temp &  0-20\\
    \midrule
 MEWS~\cite{subbe2001validation}& RR         &           &           & HR, SBP                     & GCS &              & Temp  & 0-15\\ 
 \midrule
APACHE II~\cite{knaus1985apache}& RR         & PLT, PT   &           & MAP, HR     &  GCS   & Creatinine,    & Age, Temp, PH,   & 0-71 \\
                                &            &           &           &             &     & Urine output  & SS, SP, WBC, LCT             &         \\  
    \bottomrule
    \end{tabular}
    \label{tab:multi-organ-cc}
\end{table*}

\subsection{Method Details}

\subsubsection{Model Backbone Selection}
\label{sec:model_selection}
Unlike most existing sequence-representation-based clinical prediction studies~\cite{retain,dipole,zhang2021interpretable} that treat EHRs as observational sequences, we represent patients' EHRs as graphs. 
Modeling EHRs presents several important challenges:
\begin{itemize}
\item \textbf{Temporal Sequencing of Clinical Events}: The chronological order of clinical events is crucial for accurately describing a patient's condition.
\item \textbf{Interaction of Clinical Events}: Clinical events are often closely interconnected, such as the use of vasopressors to treat extremely low mean blood pressure (MBP).
\item \textbf{High Missing Rate in Clinical Observations}: Many clinical variables, such as lab tests, often have high rates of missing data.
\end{itemize}

Sequence-based representation can handle the temporal information of EHRs well. However, the models~\cite{retain,dipole} typically require fixed-size vectors as input, which may necessitate additional operations (e.g., imputation) to address the issue of missing data in EHRs. Furthermore, the interaction between clinical events, which plays a significant role in modeling a patient's health state, is often overlooked by most sequence-based models. Failure to address these last two challenges may result in suboptimal performance of the prediction models. 

In this study, we use temporal graphs to represent patients' EHRs. Graphs can naturally model the interaction between clinical variables. Only the observed variables and estimated calculators are included in the temporal graphs, eliminating the need for additional imputation methods and avoiding potential imputation bias. Moreover, we use directed edges between clinical observation nodes to incorporate temporal information. With this graph representation, we employ a graph neural network as the model backbone to represent patients' health states and make sepsis predictions.

\subsubsection{Clinical Embedding}
\label{sec:embedding}

\textbf{Value embedding.}
For variables, we adopt value embedding~\cite{yin2020identifying,zhang2021interpretable} to map the values into vectors. Given a variable $v$ and the observed values in the whole dataset, we sort the values and discretize the values into $n (n=1000)$ sub-ranges with equal number of observed values in each sub-range. The variable $v$ is embedded into a vector $e^v \in R^d$ with an embedding layer. For the the observed value for variable $v$ within sub-range $i (1 \leq i \leq n)$, we embed it into a vector $e^{v'} \in R^{2d}$:
\vspace{-1mm}
\begin{equation}
\label{eq:value-embedding-1}
\begin{split}
e^{v'}_j = sin(\frac{i*j}{n*d})  \\
e^{v'}_{d+j} = cos(\frac{i*j}{n*d}),
\end{split}
\end{equation}
where $0 \leq j < d$. By concatenating $e^v$ and $e^{v'}$, we obtain a vector containing both the variable's and its value's information. A linear layer is followed to map the concatenation vector into a new value embedding vector $\mathbf{h}_v \in R^d$. 
\begin{equation}
\label{eq:value-embedding-2}
\mathbf{h}_v = \text{L}([e^v; e^{v'}]), 
\end{equation}
where L($\cdot$) denotes a linear mapping function. 

\textbf{Time Embedding.} In order to incorporate the elapsed time between observed values, we leverage a time embedding~\cite{yin2020identifying,zhang2021interpretable} for the time gap $\delta$:
\begin{equation} 
\begin{split}
e^{\delta}_j = sin(\frac{\delta*j}{T_m*d})   \\
e^{\delta}_{d+j} = cos(\frac{\delta*j}{T_m*d}) ,  
\end{split}
\end{equation}
where $0 \leq j < d$, $T_m$ denotes the maximum of time gap ($0 < \delta \leq T_m$). We combine the edge embedding $e^r$ with the time embedding $e^\delta$ to generate $\mathbf{h}_r \in R^d$:
\vspace{-1mm}
\begin{equation}
\mathbf{h}_r = \text{L}([e^r; e^\delta])
\vspace{-1mm}
\end{equation}



\subsection{Experiment Details}

\subsubsection{Variables Used for Sepsis Prediction}
\label{sec:variables}
Following \cite{aiclinician,yin2020identifying}, we use following variables to model sepsis patients' health states:
 heart rate, Respratory, Temperature, Spo2, SysBP, DiasBP, MeanBP, Glucose, Bicarbonate, WBC, Bands, C-Reactive, BUN, GCS, Urineoutput, Creatinine, Platelet, Sodium, Hemoglobin, Chloride, Lactate, INR, PTT, Magnesium, Aniongap, Hematocrit, PT, PaO2, SaO2, Bilirubin.
The first 8 variables are immediately available vital signs. 

\subsubsection{Missing Rates of Variables}

\autoref{tab:all-variables} displays the missing rates of the lab tests and calculators. Note that lab tests are usually observed once from several hours to days, so we display the 4-hour missing rates here. 
A missing value means the variable has not been observed for more than 4 hours.

\begin{table} [!htb]  
\setlength{\tabcolsep}{2.2pt}
\centering
\caption{Missing rates of observed lab tests and multi-organ clinical calculators related to sepsis. }  
\label{tab:all-variables} 
\begin{tabular}{cccc}
\toprule
variable & AmsterdamUMCdb & OSUWMC & MIMIC-III\\
\midrule
GCS                & 29\% & 50\% & 33\%\\
\midrule
Urine output        & 23\% & 39\% & 33\%\\
Creatinine (CRT)   & 75\% & 85\% & 80\%\\
\midrule
Platelet (PLT)     & 76\% & 88\% & 82\%\\
PTT                & 76\% & 83\% & 79\%\\
PT                 & 78\% & 92\% & 80\%\\
INR                & 78\% & 84\% & 80\%\\
\midrule
Bilirubin          & 92\% & 94\% & 93\%\\
Glucose (GLC)      & 34\% & 49\% & 36\%\\
\midrule
PaO2               & 86\% & 92\% & 87\%\\
SaO2               & 88\% & 94\% & 89\%\\
Hemoglobin (HMG)   & 56\% & 75\% & 69\%\\
Bicarbonate (BCB)  & 69\% & 74\% & 67\%\\
Lactate (LCT)      & 88\% & 90\% & 89\%\\
\midrule
WBC                & 67\% & 78\% & 69\%\\
BUN                & 63\% & 76\% & 66\%\\
Bands              & 99\% & 99\% & 99\%\\
C-reactive         & 99\% & 99\% & 99\% \\
\midrule
Magnesium          & 66\% & 76\% & 69\%\\
Aniongap (AG)      & 62\% & 78\% & 67\%\\
Hematocrit (HMT)   & 60\% & 76\% & 64\%\\
Chloride (CLR)     & 62\% & 70\% & 66\%\\
Sodium (SDM)       & 55\% & 72\% & 65\%\\
\midrule
SOFA & 94\% & 95\% & 94\% \\
APACHE II & 85\% & 92\% & 88\%\\
SIRS & 75\% & 85\% & 77\%\\

NEWS & 34\% & 55\% & 36\%\\
MEWS & 33\% & 54\% & 36\%\\
qSOFA & 33\% & 53\% & 35\%\\
\bottomrule
\end{tabular}  
\end{table}

\subsubsection{Methods for Comparison}
\label{sec:model_comparison}

To validate the performance of the proposed framework for early sepsis risk prediction task, we compare the propose \ours~ to the following models: 

\begin{itemize}
    \item Clinical calculator-based methods: We use the widely used clinical calculators (i.e., \textbf{NEWS}~\cite{smith2013ability}, \textbf{MEWS}~\cite{subbe2001validation}, \textbf{qSOFA}~\cite{sepsis3}, \textbf{SIRS}~\cite{bone1992definitions}) to build the sepsis prediction models. A logistic regression is used to predict sepsis with the calculator scores,  the component variables, and frequently observed vital signs.

    

    \item \textbf{GRU} and \textbf{LSTM}: GRU \cite{gru} and LSTM \cite{lstm} are classical RNN based models, which both introduce various gates to improve RNN's performance. 

    \item \textbf{RETAIN:} The REverse Time AttentIoN model (RETAIN) \cite{retain} is the first work that tries to interpretate model's disease risk prediction results with two attention modules. The attention modules generate weights for every medical events. The weights are helpful to analyze different events' contributions to the output risk.
    \item \textbf{Dipole}~\cite{dipole}: Dipole employs bidirectional recurrent neural networks combined with three distinct attention mechanisms for patient visit information prediction.

    \item  \textbf{DFSP} \cite{duan2023early}: Double Fusion Sepsis Predictor (DFSP) is an early sepsis prediction model that uses early and late fusion techniques to improve the accuracy and robustness of sepsis prediction.

    \item \textbf{RGNN}~\cite{liu2020hybrid}: RGNN is a hybrid method of RNN and GNN with RNN to represent patient status sequences and GNN to represent temporal medical event graphs like \autoref{fig:graph_construction}(B).

    \item \textbf{GTN}~\cite{chowdhury2023predicting}: GTN also represent EHRs as graphs and adopt a Transformer~\cite{vaswani2017attention} to make clinical risk predictions.

\end{itemize}

\subsubsection{Implement Details}
\label{sec:implementaion}

We implement our proposed model with Python 3.8.10 and PyTorch 1.12.1\footnote{\url{https://pytorch.org/}}. For training models, we use Adam optimizer with a mini-batch of 64 patients. The multi-modal data are projected into a $512$-d space ($d = 512$). We train the proposed model on 1 GPU (TITAN RTX 6000), with a learning rate of 0.001. We randomly divide the patients in datasets into 10 sets. All the experiment results are averaged from 10-fold cross-validation, in which 7 sets are used for training every time, 1 set for validation, and 2 sets for testing. The validation sets are used to determine the best values of parameters in the training iterations. We ran the training and test phases 10 times and reported the mean and standard deviation of the metrics in \autoref{sec:result}.

For non-graph-based models, we normalize the values of variable $i$ as follows:
\begin{equation}
\label{eq:normalization}
x^i = \frac{x^i - mean(x^i)}{std(x^i)} ,
\end{equation}
where $mean$ and $std$ are the mean value and standard deviation for the variable $i$ on the whole dataset. 
Because the non-graph-based models cannot handle missing variables, we use a popular imputation method 3D-MICE~\cite{3dmice} to impute the missing values.

\subsubsection{Evaluation Metrics}

For the sepsis prediction tasks, we use Area Under the Receiver Operating Characteristic Curve (AUC), F1, and Recall for evaluation metrics.
For the calculator estimation tasks,  we measure the models' performance  with nRMSE. The nRMSE is calculated from the gap between the ground truth and prediction.
Given a variable $i$, nRMSE is defined as:
\vspace{-1mm}
\begin{equation}
\label{eq:nmse}
nRMSE^{i} = \sqrt{\frac{\sum_j \sum_t a^{(j),i}_t (\widetilde{x}^{(j),i}_t - \hat{x}^{(j),i}_t)^2 }{\sum_j \sum_t a^{(j),i}_t}} , 
\end{equation}
where $\hat{x}^{(j),i}_t$, $\widetilde{x}^{(j),i}_t$, $a^{(j),i}_t$ indicate the ground truth, imputed value, and masking indicator for patient $j$, variable $i$ in collection $t$.  

\subsubsection{Clinical Event Interaction}

As \autoref{fig:graph_construction}(C) shows, we incorporate the clinical event interaction to build the temporal graph.  
Following the surviving sepsis campaign bundle~\cite{levy2018surviving}, 
we consider two kinds of important treatments for septic patients:
vasopressors and IV fluid to prevent low blood pressure (related variables: SBP, DBP, DBP), antibiotics to treat infections (related variables: WBC, BUN, Bands, C-reactive). 
We also consider mechanical ventilation as the treatment for acute respiratory distress syndrome (related variables: SpO2, PaO2, SaO2, respiratory rate), which frequently co-occurs with sepsis~\cite{SANCHEZ2023135}.
We add the interaction relation between the treatments and related variables to the constructed temporal graph for patient health status modeling.

\subsection{Additional Experiments}

\subsubsection{Organ Dysfunction Prediction}

This study aims to early identify the patients with potential risk. We adopt additional prediction branches to force the model to learn the organ dysfunction patterns with $\mathcal{L}_o$ in \autoref{eq:bceloss}. 

\noindent
\textbf{Organ Dysfunction Prediction Setting.} We use the same setting as sepsis prediction (to predict whether the specific organs will suffer from dysfunction with a 4-hour sliding window, similar to \autoref{fig:setting}) to predict organ dysfunction risk.

The results of organ dysfunction predictions are presented in \autoref{tab:organ_dysfunction_prediction}. The findings show that \ours~ outperforms the other versions (\textit{i.e.}, \ours$^{imp}$, \ours$^{-i}$, \ours$^{-d}$). The results demonstrate the effectiveness of the proposed graph construction module in organ dysfunction prediction tasks, which could further improve the sepsis prediction performance of \ours~ in \autoref{tab:missing_cal_sepsis_prediction}.

\begin{table}[]
    \centering
    \setlength{\tabcolsep}{1.2pt}
    \caption{Organ dysfunction prediction results.}
    \vspace{-2mm}
    \begin{tabular}{l|ccc|ccc|ccc}
    \toprule
        Method & \multicolumn{3}{c|}{MIMIC-III} & \multicolumn{3}{c|}{\small AmsterdamUMCdb} & \multicolumn{3}{c}{OSUWMC}\\ 
        & AUC & F1 & \small{Recall} & AUC &  F1 & \small{Recall} & AUC & F1  & \small{Recall} \\
        \midrule
        \small \ours$^{imp}$ & .835 & .440 & .761 & .840 & .443 & .778 & .912 & .470 & .839 \\
        \small \ours$^{-i}$ & .832 & .439 & .758 & .831 & .445 & .775 & .910 & .468 & .832 \\
        \small \ours$^{-d}$ & .841 & .445 & .767 & .842 & .445 & .780 & .912 & .468 & .835 \\
         \small \ours       & \textbf{.862} & \textbf{.458} & \textbf{.787} & \textbf{.865} & \textbf{.453} & \textbf{.790} & \textbf{.925} & \textbf{.485} & \textbf{.856} \\
        \bottomrule
    \end{tabular}
    \label{tab:organ_dysfunction_prediction}
\end{table}

\subsubsection{Effectiveness of Clinical Calculators}

When all the component variables are observed, the ground truths of the clinical calculators are available, we use nRMSE to evaluate the calculator estimation performance. \autoref{tab:nrmse_gt} shows that the nRMSE between the ground truths and the estimated calculators is close to 0, demonstrating that our model can accurately learn the computation mechanisms of clinical calculators.




\begin{table}[]
    \centering
    \setlength{\tabcolsep}{2pt}
    \caption{nRMSE of calculator estimation (full observation setting). All the component variables of the calculators are observed and the ground truths of calculators are available.}
    \begin{tabular}{l|ccccccc}
    \toprule
         Dataset & SOFA & \small{APACHEII} & qSOFA & SIRS & MEWS & NEWS\\
        \midrule
         MIMIC-III & 0.01 & 0.01 & 0.01 & 0.01 & 0.01 & 0.01\\
         \small AmsterdamUMCdb  & 0.01 & 0.02 &  0.01 & 0.01 & 0.02 & 0.01 \\
         OSUWMC  & 0.01 & 0.02 & 0.01 & 0.01 &  0.01 & 0.01\\ 
         \bottomrule
    \end{tabular}
    \label{tab:nrmse_gt}
\end{table}

\subsubsection{Hyper-Parameter Optimization}
We use the sum of four loss functions to train the model in \autoref{eq:sum_loss}. 
We also tried to adjust the weights $\alpha_o, \alpha_e, \alpha_d$ for the loss functions. We conducted a grid search to find the best $\alpha_*$ with the following values [0.1, 0.2, 0.3, 0.5, 1, 2, 3, 5, 10]. We found the models achieved the best performance with $0.3 \le \alpha_* \le 3$ and are not sensitive to the weights, so we set $\alpha_* = 1$ for the proposed \ours.

\subsection{Additional Details for Deployment}
\begin{figure}
    \centering
    \includegraphics[width=\linewidth]{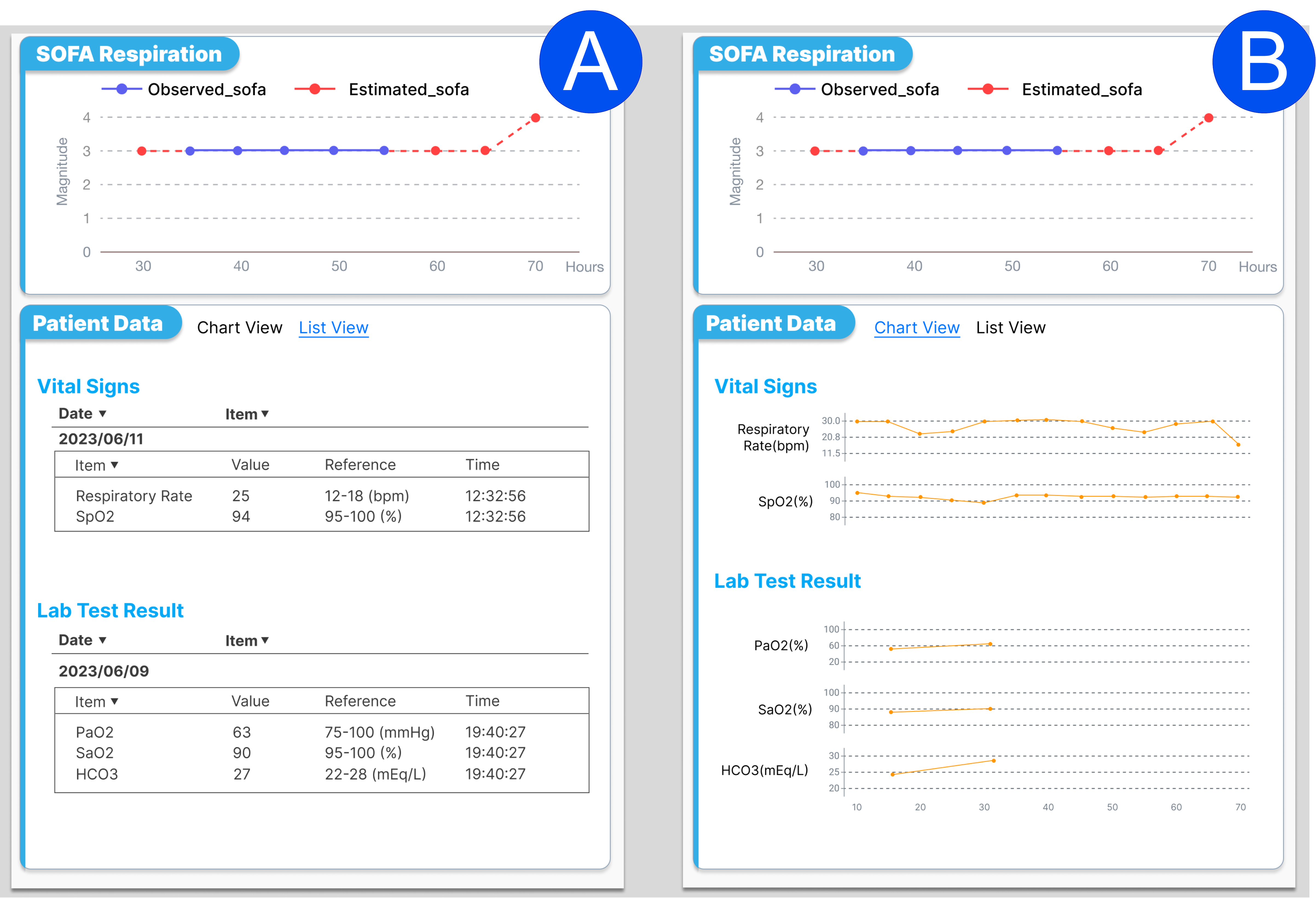}
    \vspace{-4mm}
    \caption{(A) List view of clinical variables for organ status. (B) Chart view of clinical variables for organ status. }
    \label{fig:detail-UI}
\end{figure}

Note that our \ours~ system offers both list view (\autoref{fig:detail-UI}(A)) and chart view (\autoref{fig:detail-UI}(B)) to display the sequences of observed variables and the latest observed values with reference ranges, such that the clinicians can track the relative change trends and absolute dysfunction status for various organ systems.

Note that clinicians can interact with the system to display the dysfunction status of various organ systems. \autoref{fig:detail-UI} just displays the variables related to the respiratory system for illustration purposes.

\end{document}